\documentclass[aip,jcp,floatfix,
 amsmath,amssymb,
 reprint,
 colorinlistoftodos,
]{revtex4-2}

\usepackage{graphicx}
\usepackage{dcolumn}
\usepackage{bm}

\usepackage[utf8]{inputenc}
\usepackage[T1]{fontenc}
\usepackage{mathptmx}

\usepackage{amsmath}
\usepackage{multirow}
\usepackage{amsfonts}

\usepackage[draft]{changes}

\usepackage{etoolbox}
\usepackage{soul}
\usepackage{booktabs}

\usepackage{todonotes}

\makeatletter
\def\@email#1#2{%
 \endgroup
 \patchcmd{\titleblock@produce}
  {\frontmatter@RRAPformat}
  {\frontmatter@RRAPformat{\produce@RRAP{*#1\href{mailto:#2}{#2}}}\frontmatter@RRAPformat}
  {}{}
}%
\makeatother

\preprint{AIP/123-QED}

\def\BibTeX{{\rm B\kern-.05em{\sc i\kern-.025em b}\kern-.08em
    T\kern-.1667em\lower.7ex\hbox{E}\kern-.125emX}}

\begin{document}

\title{Generative Image Restoration and Super-Resolution using Physics-Informed Synthetic Data for Scanning Tunneling Microscopy}

\author{Nikola L. Kolev}
\email{nikola.kolev.21@ucl.ac.uk.}
\altaffiliation[]{These authors have contributed equally to this work.}
\affiliation{London Centre for Nanotechnology, University College London, London, United Kingdom}
\affiliation{Department of Electronic and Electrical Engineering,
University College London, London, United Kingdom}

\author{Tommaso Rodani}
\email{tommaso.rodani@areasciencepark.it}
\altaffiliation[]{These authors have contributed equally to this work.}
\affiliation{AREA Science Park, Trieste, Italy}
\affiliation{University of Trieste, Trieste, Italy}

\author{Neil J. Curson}
\affiliation{London Centre for Nanotechnology, University College London, London, United Kingdom}
\affiliation{Department of Electronic and Electrical Engineering,
University College London, London, United Kingdom}

\author{Taylor J.Z. Stock}
\altaffiliation[]{These authors have contributed equally to the supervision of this work.}
\affiliation{London Centre for Nanotechnology, University College London, London, United Kingdom}
\affiliation{Department of Electronic and Electrical Engineering,
University College London, London, United Kingdom}

\author{Alberto Cazzaniga}
\altaffiliation[]{These authors have contributed equally to the supervision of this work.}
\affiliation{AREA Science Park, Trieste, Italy}

\date{29 Oct 2025}

\begin{abstract}
Scanning tunnelling microscopy (STM) enables atomic-resolution imaging and atom manipulation, but its utility is often limited by tip degradation and slow serial data acquisition. Fabrication adds another layer of complexity since the tip is often subjected to large voltages, which may alter the shape of its apex, requiring it to be conditioned. Here, we propose a machine learning (ML) approach for image repair and super-resolution to alleviate both challenges. Using a dataset of only 36 pristine experimental images of Si(001):H, we demonstrate that a physics-informed synthetic data generation pipeline can be used to train several state-of-the-art flow-matching and diffusion models.  Quantitative evaluation with metrics such as the CLIP Maximum Mean Discrepancy (CMMD) score and structural similarity demonstrates that our models are able to effectively restore images and offer a two- to fourfold reduction in image acquisition time by accurately reconstructing images from sparsely sampled data. Our framework has the potential to significantly increase STM experimental throughput by offering a route to reducing the frequency of tip-conditioning procedures and  to enhancing frame rates in existing high-speed STM systems.
\end{abstract}

\maketitle

\section{Introduction}
Scanning tunnelling microscopy (STM), along with related techniques such as atomic-resolution atomic force microscopy (AFM), has been widely adopted for imaging and manipulation at atomic scales since its inception in 1981 \cite{binnigreconstruction} and has made possible a range of studies on the nanoscale. Despite significant advancements, STM still struggles with two primary bottlenecks: the slow rate of image acquisition and the frequent degradation of the tip's quality. 

Tip contamination or mechanical deformations can severely degrade imaging quality and attaining stable atomic resolution typically demands extensive manual tip conditioning by highly skilled operators. It also lacks robust automated solutions across varied materials and experimental conditions. 
Similarly, STM imaging can be orders of magnitude slower compared to methods such as scanning transmission electron microscopy (STEM). Video frame rate STM exists, but requires specialised hardware \cite{rost2005scanning}. The high tip speeds needed mean constant height mode must be used. This leads to lower image quality and means the surfaces imaged must be relatively flat \cite{rost2005scanning}. These limitations significantly hinder the efficiency and scalability of STM, especially in applications requiring high throughput.

Importantly, our goal is not to take arbitrary or extremely degraded STM images and reconstruct them at atomic resolution - such an approach would run the risk of inaccuracies due to hallucinated features. Instead, our method is most useful for tasks where the surface is well understood or speed and automation is preferable over perfect accuracy, such as STM applications in hydrogen desorption lithography (HDL), atomic manipulation, or sample navigation \cite{fricke2021coherent, ramsauer2023autonomous, leinen2020autonomous}. 

Improving a degraded tip state has been the focus of a number of studies \cite{ gordon2020embedding, rashidi2018autonomous, barker2024automated, krull2020artificial, zhu2024autonomous} and is the prominent direction within the STM community to improve data collection via ML-assistance. One such framework employed reinforcement learning (RL) to select among six predefined actions, an improvement over random choice as used by Barker et al. \cite{barker2024automated} and Rashidi et al. \cite{rashidi2018autonomous}, but still a limited approximation of the wide range of corrective options available to an STM user. Nevertheless, reinforcement learning enabled a significant speed-up - approximately four to fivefold - compared to random action selection, with improved consistency too. Gordon et al. \cite{gordon2020embedding} went some way towards embedding human evaluation strategies into their tip evaluation tool. 

Their system used multiple classification categories rather than a simple binary scheme and incorporated tip-condition assessment during scanning rather than relying on a complete image. Specifically, they used one network for recent scan lines and a long short-term memory (LSTM) network to integrate historical data, producing a more context-aware evaluation. However, this richer analysis required a substantially larger data set of 6,167 images. 

In contrast, other works proposed methods that reduce the demand for data by semi-supervised labelling \cite{rodani2023towards} or synthetic data generation \cite{rodani24enhancing}. Barker et al. \cite{barker2024automated} uses an autocorrelation function to classify tip states, their approach required only a single reference image to evaluate new data. Although this technique depends on the presence of a distinctive repeating surface feature - an assumption that may not always hold - the dramatic reduction in dataset requirements is noteworthy. Moreover, such an approach could potentially be combined with more complex frameworks, such as Gordon et al.’s, by generating labelled data at low cost.

In this study, we introduce a complementary computational strategy that directly corrects image artefacts. Our core contribution is a physics-informed framework that learns to restore images and perform super resolution using only a minimal set of pristine images. This approach reduces the frequency of tip-conditioning cycles and shortens experimental runtimes. The reduction in conditioning not only decreases the time spent assessing tip quality but also minimises the need to navigate between different surface areas. This is particularly advantageous in applications where specific regions must be revisited - such as in atomically precise fabrication - since, without correction, piezoelectric hysteresis prevents the same nominal $(x, y)$ coordinates from corresponding to the same physical location after large tip movements.

We also examine ML methods to reduce scan duration by super-resolving low resolution images. In general, for STM to be a viable fabrication method outside the laboratory, it must operate faster - current state-of-the-art HDL requires many hours to produce a single electron transistor \cite{KspruceThesis}. In addition, attempts to use RL to automate atomic manipulation suggest that the ever changing tip apex requires an RL agent to be retrained when drastic tip changes occur \cite{leinen2020autonomous}. Reducing the probability of a tip change by shortening tip-sample interactions (e.g., by taking lower resolution images and thus minimising the tip's path) can extend the agent's useful operational time. 

We present a deep-learning approach tailored to overcome these persistent STM challenges through advanced image restoration and targeted super-resolution (SR). Central to our approach is the augmentation of experimental STM images with carefully modelled artefacts such as multi-tip, scan line noise, and tip blunting. Because this strategy relies on physical modeling of these artefacts, unlike traditional techniques that often use simple geometric transformations, it yields training data that better captures the variability of real-world experimental conditions. In contrast, earlier studies \cite{joucken2022denoising, xie2024physics} relied on simplified noise models. For example, Joucken et al. introduced Gaussian noise and scan line misalignments but neglected critical artefacts such as multi-tip distortions and tip blurring.

We systematically evaluated multiple generative models, focussing on flow-matching (FM) \cite{lipman2022flow} and denoising diffusion implicit models (DDIM) \cite{song2020denoising} due to their ease of training, rapid inference compared to denoising diffusion probabilistic models (DDPM), and superior performance \cite{rombach2022high}. Recent explorations into the use of ML for microscopy image repair have focused largely on general adversarial networks (GANs) or autoencoders \cite{joucken2022denoising, xie2024physics, khan2023leveraging}. Although GANs show excellent results, their training is far from simple and is known to be unstable and difficult to converge \cite{saad2024survey}.

Understanding practical constraints encountered in typical STM laboratories, we further examined model performance differences between GPU and CPU environments.

Our results highlight that the FM approach significantly outperforms a traditional autoencoder. Importantly, our models efficiently remove severe tip-induced artefacts while preserving essential atomic-scale details when restoring images. Furthermore, the proposed SR approach successfully accelerates STM image acquisition rates by 2–4 times, reconstructing sparsely sampled 512$\times$512 pixels images in approximately 11 seconds on standard CPUs.

\section{Method}
Training neural networks requires large datasets, which are uncommon in experimental fields such as STM. To address this, we developed a physics-informed synthetic data generation pipeline that applies realistic instrumental artefacts to high-quality experimental data. The pipeline builds upon a small experimental dataset of 54 pristine images of the Si(001):H surface, each with dimensions of 512$\times$512 pixels (corresponding to 100~nm$\times$100~nm). These were divided into training (36 images), validation (12 images), and test (6 images) subsets. To evaluate the generalisation of the models on real-world data we curated a set of 66  512$\times$512 pixels (100~nm$\times$100~nm) degraded experimental images for the task of image restoration. For the super-resolution task, we collected images at multiple resolutions: 4 at 512$\times$256 pixels (100~nm$\times$100~nm), 3 at 256$\times$128 pixels images (50~nm$\times$50~nm), 4 at 512$\times$128 pixels (100~nm$\times$100~nm), and 3 at 256$\times$64 pixels images (50~nm$\times$50~nm).

The syntethic data generation process begins by choosing a random image from a training, validation, or test set and normalising it between 0 and 1, after which a sequence of stochastic transformations was applied to simulate common experimental degradations and increase dataset size, which we outline in Section~\ref{sec:data_augmentation}.

The synthesis pipeline produced three datasets for the three tasks we perform using these initial splits: image restoration, super-resolution (SR) of a $2\times$, and SR of $4\times$. Each of these has a training set of $20,000$ samples, a validation set of $2,000$, and a synthetic test set of $2,000$. Each sample is a two-channel tensor containing the pristine, high-resolution ground truth and the synthetically degraded image. 

In order to train a more robust SR model that works with imperfect data, we also synthetically degrade SR proxies. This means that the models we train for SR have to perform both restoration and SR. Section~\ref{sec:data_augmentation} outlines our sampling procedure when constructing the SR datasets, as well as the data augmentation strategies used to improve model generalisation.

Furthermore, to isolate performance on specific degradation types, we generated targeted test sets of $1,000$ samples each for multi-tip, scan line misalignment, tip change, and blunt tip artefacts, as shown in Figure~\ref{fig:noise}(f–j). Equivalent sets were also created for the SR task, along with a low-resolution, degradation-free baseline set. These datasets collectively provide a controlled and diverse foundation for evaluating the effectiveness of our augmentation and training strategies, and identify degradation types are most difficult to restore.

\subsection{Synthetic Data Generation} \label{sec:data_augmentation}

\begin{figure}[htbp]
    \hspace*{-7mm}
    \makebox[\linewidth][l]{
    \includegraphics[width=0.4\textwidth]{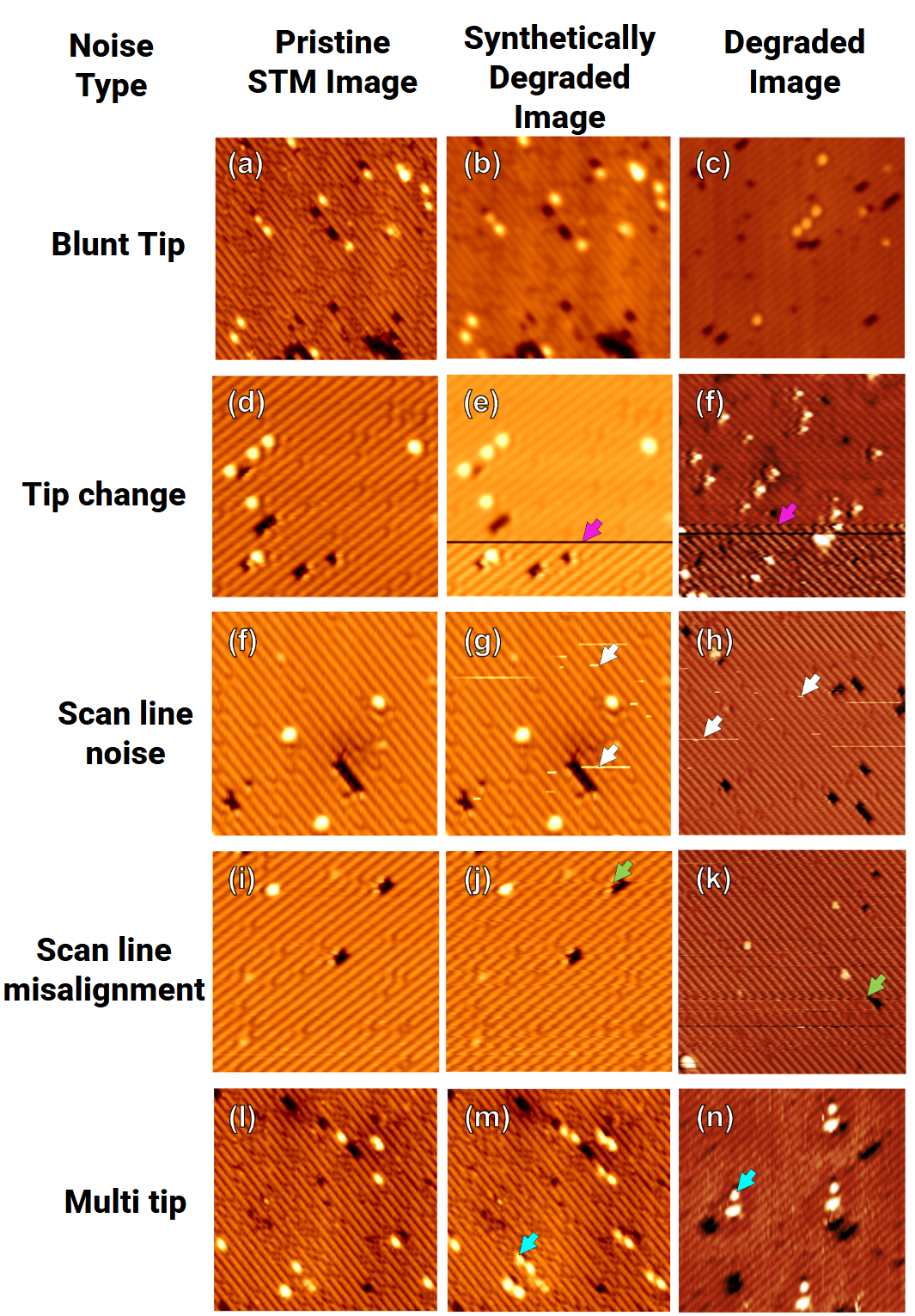}
    }
    \caption{Comparison of experimental and synthetic STM degradations. Each image is 25~nm$\times$25~nm (taken at -2~V and between 20~pA and 60~pA). The left column are pristine experimental images, and the middle column are the synthetically degraded pristine images. The right column are degraded experimental images. Each row shows a specific degradation type that that we later aim to correct. }
    \label{fig:noise}
\end{figure}

The synthetic data generation employed in this work is central to our method. It allows us to avoid labelling large datasets, requiring only a small curated set of pristine experimental images, while still exploiting the benefits of ML algorithms. Each image degradation type was parametrised by random variables drawn from prescribed distributions, ensuring a varied and realistic dataset. In general, distribution ranges were selected by quantitative comparison with experimental noise (e.g., the height and shape of scan line noise) or by visual inspection (e.g., the $\sigma$ used for blunt-tip blurring). In addition, the expression used for the multi-tip artefacts was motivated via a quantum mechanical derivation, details of which are given in Appendix~\ref{sec: Double_tip_derivation}. As well as the four main degradation types shown in Figure~\ref{fig:noise}, we applied random rotations and crops to further increase our dataset size. The full generation pipeline, in order of application, is as follows, with the degradation types in bold:

\begin{itemize}
    \item (1) Random Rotation: Each image was rotated by 0, 90, 180, or 270 degrees with equal probability.
    
    \item \textbf{(2) Multi-tip Artefacts:} Multi-tip effects were simulated by superimposing up to four displaced copies of the clean image, $h(x, y)$, applied with a probability of 0.5 for 2 tips, 0.3 for 3 tips, and 0.2 for 4 tips. Each copy was modified by a sigmoid function, randomly offset and added back on to the original image
    \begin{equation} \label{eqn: double_tip_eqn}
    f(x,y) = h(x, y) + \sum\limits_{i=1}^{N}K_i\Bigg(\frac{A_i}{1 + e^{\,c_i - d_ih(x-\tilde{x}_i, y-\tilde{y}_i)}}\Bigg)
    \end{equation}
    where \(N \sim \mathrm{Cat}(\{2,3,4\}; \tfrac12,\tfrac{3}{10},\tfrac15)\), \(c_i \sim \mathcal{U}(5,9)\), \(d_i \sim \mathcal{U}(7,10)\), amplitude multiplier \(A_i \sim \mathcal{U}(1,2.5)\), and random offsets of \(\tilde{x}_i, \tilde{y}_i \sim \mathcal{U}(1,11)\). $K_i(\cdot)$ is a kernel applied to the doubled image to simulate a different tip shape from the original and it is selected probabilistically from:  
    \begin{itemize}
        \item Gaussian filter with standard deviation \(\sigma \sim \mathcal{U}(1,3)\), with probability 0.3;  
        \item Median filter with kernel size \(k \in \{1, \dots, 9\}\), with probability 0.4;  
        \item Random filter with entries drawn from \(\mathcal{U}(-0.5,1)\) and kernel size \(k \in \{5,6\}\), with probability 0.3.  
    \end{itemize}
    
    \item \textbf{(3) Scan line Misalignment:} Horizontal misalignments were introduced with probability 0.3 by shifting in$x$ individual scan lines by an amount drawn from a Gaussian distribution with \(\sigma = 0.8\).
    
    \item (4) Random Crop: A random crop of \(128 \times 128\) pixels was extracted.
    
    \item \textbf{(5) Blunt tip:} To simulate the blurring effect of a blunt tip, a Gaussian filter with \(\sigma \sim \mathcal{U}(0.3,0.6)\) was applied with probability 0.6.
    
    \item \textbf{(6) Tip change:} Abrupt changes in the tip apex were simulated by blurring the image from a randomly chosen scan line onwards with probability 0.6. This mimics a sudden degradation in instrument resolution. With probability 0.5, a constant offset, $\Delta$, was also added to the scan line where the blurring begins, $\Delta \sim \mathcal{U}(s \times 0.05, \, s \times 0.4), \quad s \sim \mathrm{Unif}(\{-1,1\}).$
    
    \item (7) Downsample and Upsample: For SR datasets, downsampling was applied at this stage to avoid interfering with the scan line noise in step 8. Images were downsampled by factors of \(4\times\) and \(2\times\) in the \(y\)-direction only, then upsampled back to their original size using nearest-neighbour interpolation. This one-dimensional downsampling reflects the raster nature of STMs, where the imaging time depends on the number of scan lines and the tip speed. The number of pixels per line does not affect the total length of the scan path. The tip speed should remain constant, since increasing it can degrade image quality by exceeding the feedback loop’s response time. By reducing the number of scan lines, the total acquisition time decreases, while the pixel density along each line - and thus the in-line spatial resolution - is preserved. More information on STM raster imaging can be found in Appendix~\ref{sec:STM_background}.
    
    \item \textbf{(8) Scan line Noise:} Scan line noise was introduced with probability 0.6. The number of affected lines, \(m\), was sampled from \(\mathcal{U}(25,35)\). Noise segments had lengths drawn from \(\mathcal{U}(0,102)\), and each segment was perturbed by one of three functions:  
    \begin{itemize}
        \item Constant offset from \(\mathcal{U}(0,0.4)\), with probability 0.3;  
        \item Log-normal function \(\mathcal{LN}(\mu,\sigma)\), with \(\mu \sim \mathcal{U}(1,2)\), \(\sigma \sim \mathcal{U}(0.5,1)\), with probability 0.45, representing a sudden tip jump with gradual recovery;  
        \item Sinusoidal function, with probability 0.25.  
    \end{itemize}
    \item (9) Normalisation: Each of the channels is independently normalised to a range of $[-1,1]$.
\end{itemize}

Images were cropped to $128 \times 128$ to achieve a larger total effective area. Larger images were processed by dividing them into overlapping patches, which were individually restored or upsampled, and then recombined. A small overlap between patches, combined with the application of a squared-cosine window, ensures smoother transitions and minimises visible stitching artefacts.

\subsection{Training Objectives and Architectures} \label{sec:training_obj}

Our approach to image restoration is centred on conditioned generative models built upon the U-Net architecture. We investigate the trade-off between model capacity and computational cost by designing two distinct configurations: a compact, parameter-efficient model (3 blocks, no attention, channel capacities of $[32, 64, 128]$) and a larger, higher-capacity variant (3 blocks, central self-attention layer, channel capacities of $[64, 128, 256]$). We give a summary of the different model variants and their names in Table~\ref{tab:model_variants}. 

\begin{table}[htbp!]
    \centering
    \caption{Summary of the model variants tested in this study. The models differ in their architectural capacity, determined by channel dimensions and the inclusion of a self-attention layer. Each model is trained with a specific loss function corresponding to its framework: Flow-Matching ($L_{FM}$), standard DDIM ($L_{DM}$), DDIM with an auxiliary Fourier-transform loss ($L_{FT,DM}$), and a mean absolute error loss for the autoencoder baseline ($L_{MAE}$). The size of the models is given in millions (M) of trainable parameters. }
    \label{tab:model_variants}
    \begin{tabular}{lccc}
        \toprule
        \multirow{2}{*}{\textbf{Model}} & \multirow{2}{*}{\textbf{Parameters (M)}} &  \textbf{Self-attention}  &\multirow{2}{*}{\textbf{Loss}} \\
         & & \textbf{layer} & \\
        \midrule
         Autoencoder   & 14.7 &\checkmark   & $L_{MAE}$ \\
         DDIM Small     & 3.6  & $\times$     & $L_{DM}$ \\
         DDIM Large  & 14.7 & \checkmark   & $L_{DM}$ , $L_{FT,DM}$ \\
         FM Small       & 3.6  & $\times$     & $L_{FM}$ \\
         FM Large       & 14.7 & \checkmark   & $L_{FM}$ \\
        \bottomrule
    \end{tabular}
\end{table}

Both architectures were trained for denoising and super-resolution (SR) tasks ($2\times$ and $4\times$ upscaling) using two state-of-the-art generative frameworks: Flow-Matching (FM) and Denoising Diffusion Implicit Models (DDIM). Additionally, we add an FFT-based loss to a subset of DDIM models. This choice is motivated by previous work showing that Fourier-domain losses counter the spatial low frequency bias of Mean Squared Error (MSE) and improve high frequency detail and edge fidelity in SR and image restoration tasks\cite{fuoli2021fourier, cho2021rethinking, shang2024resdiff}. For comparison, we also trained two baseline models previously applied in the STM community for image restoration: an autoencoder \cite{joucken2022denoising} and a GAN specifically designed for STM images \cite{xie2024physics}. Although the autoencoder trained successfully, we were unable to reproduce stable convergence of the GAN. As the original code and detailed architectural specifications were not available, a further implementation was not feasible. We note that this outcome may reflect differences in the experimental setup rather than in the approach itself, and further motivated our focus on architectures with easier convergence.

Both FM and DDIM share the central idea of defining a forward process in which Gaussian noise is gradually added to a data sample until it becomes indistinguishable from random noise. A neural network is then trained to approximate the reverse process. After training, the model can generate new data starting from random noise and applying the learnt reverse dynamics, progressively reconstructing an image. By conditioning the model with auxiliary information (e.g., a low-resolution input), this mechanism can be adapted to tasks such as denoising, super-resolution, and inpainting \cite{song2020denoising, liu2025erase}.

Formally, each $N\times N$ image can be represented as a vector $\vec{x}_0$ in a $M$-dimensional space, where $M=N \times N$. The distribution of pristine STM images is then defined as $q(\vec{x}_0)$, which our generative models aim to approximate. In what follows, we outline the details of both approaches, including the specific noising procedures, loss functions, and inference methods.

\subsubsection*{Diffusion models (DDIM)}
In DDIM, the forward diffusion process gradually adds Gaussian noise to the data in $T$ discrete steps, yielding latent variables $\vec{x}_1, \vec{x}_2, \ldots, \vec{x}_T$. Each step is defined by
\begin{equation} \label{eq:ddpm_forward}
    \vec{x}_t = \sqrt{\alpha_t}\vec{x}_0 + \sqrt{1-\alpha_t}\vec{\epsilon}, \quad \vec{\epsilon} \sim \mathcal{N}(\vec{0}, I),
\end{equation}
where $\alpha_t$ is a decreasing sequence from 1 ($t=0$) to 0 ($t=T$, with $T \approx 1000$ in typical implementations). We use the popular implementation of Nichol et al. \cite{pmlr-v139-nichol21a}. The forward process is fixed and requires no learning. 

The neural network, $\vec{\epsilon}_{\theta}$, predicts the added noise, and the training objective is
\begin{equation}
    L_{DM} = \mathbb{E}_{t, \vec{x}_0, \vec{\epsilon}} \big[ \lVert \vec{\epsilon} - \vec{\epsilon}_\theta(\vec{x}_t, t) \rVert \big].
\end{equation}
The Fourier transform of STM images contains complementary structural information since the lattice periodicity can be captured more easily in reciprocal space. An FFT loss would therefore complement the real space domain training:
\begin{equation}
\begin{split}
    L_{FT,DM} &= \tfrac{1}{2} L_{DM}
    + \tfrac{1}{4}\,\mathbb{E}_{t, \vec{x}_0, \vec{\epsilon}} \big[\lVert |F(\vec{x}_0)| - |F(\vec{x}_{t,\theta})| \rVert \big] \\
    &\quad + \tfrac{1}{4}\,\mathbb{E}_{t, \vec{x}_0, \vec{\epsilon}} \big[\lVert \arg(F(\vec{x}_0)) - \arg(F(\vec{x}_{t,\theta})) \rVert \big],
\end{split}
\end{equation}
where $F(\cdot)$ denotes the Fourier transform, and $\vec{x}_{t,\theta}$ is the reconstruction of $\vec{x}_t$ predicted via $\vec{\epsilon}_{\theta}$ and Equation~\ref{eq:ddpm_forward}.

Inference reduces to iteratively predicting less noisy states. DDIM uses $\vec{\epsilon}_\theta$ and Equation~\ref{eq:ddpm_forward} and each step is expressed as:
\begin{equation}
\begin{split}
    \vec{x}_{t-1} &= \sqrt{\alpha_{t-1}} 
    \left( \frac{\vec{x}_t - \sqrt{1-\alpha_t}\vec{\epsilon}_\theta(\vec{x}_t, t)}{\sqrt{\alpha_t}} \right) \\
    &\quad + \sqrt{1-\alpha_{t-1}}\,\vec{\epsilon}_\theta(\vec{x}_t, t).
\end{split}
\end{equation}
This is repeated until a prediction for $\vec{x}_0$ is obtained. Crucially, DDIM allows for skipping steps: any $\vec{x}_{t-n}$ can be predicted directly, which reduces inference time, though often at the expense of output fidelity.

\subsubsection*{Flow-matching (FM)}
In FM, the forward process is defined by a simpler linear interpolation:
\begin{equation} \label{eq:fm_forward}
    \vec{x}_t = \tfrac{t}{T}\vec{x}_0 + \Big(1-\tfrac{t}{T}\Big)\vec{\epsilon}, \quad \vec{\epsilon} \sim \mathcal{N}(\vec{0}, I).
\end{equation}
Unlike DDIM, which learns a time-dependent denoising function, FM aims to learn a time-dependent velocity field $\vec{v}_(\vec{x}_t, t)$ that describes the dynamics between noisy and clean data samples \cite{lipman2022flow}. The training objective is
\begin{equation}
\begin{split}
    L_{FM} &= \mathbb{E}_{t, \vec{x}_0, \vec{\epsilon}} 
    \big[ \lVert \vec{v}(\vec{x}_t, t) - \vec{v}_\theta(\vec{x}_t, t) \rVert \big] \\
    &= \mathbb{E}_{t, \vec{x}_0, \vec{\epsilon}} 
    \big[ \lVert \vec{x}_0 - \vec{\epsilon} - \vec{v}_\theta(\vec{x}_t, t) \rVert \big].
\end{split}
\end{equation}

Where $\vec{v}_\theta(\vec{x}_t, t)$ is the FM network. Inference begins from Gaussian noise, and the learnt velocity field is integrated numerically. 
We use the second-order Runge–Kutta (RK2) midpoint method, which provides greater stability and accuracy than Euler integration. 
At each step, we first evaluate the velocity at the current point, then estimate it at the midpoint, and finally update using the midpoint estimate:
\begin{equation}
\begin{split}
    \vec{k}_1 &= \vec{v}_\theta(\vec{x}_t, t), \\
    \vec{k}_2 &= \vec{v}_\theta\!\Big(\vec{x}_t + \tfrac{\Delta t}{2}\vec{k}_1, \, t - \tfrac{\Delta t}{2}\Big), \\
    \vec{x}_{t-\Delta t} &= \vec{x}_t + \Delta t \cdot \vec{k}_2.
\end{split}
\end{equation}

where $\Delta t$ is a user-defined step size, allowing a flexible trade-off between the number of inference steps and the quality of generated samples.

\subsection{Evaluation Metrics}

In both methods, the number of inference steps plays a crucial role. Increasing the number of steps typically improves reconstruction quality, but at the cost of higher computational demand and slower image reconstruction. We evaluate both frameworks across a range of inference steps (i.e. 2, 5, and 10) to characterise the balance between inference speed and reconstruction quality. The quantitative evaluation of model performance is conducted using two categories of metrics. For synthetic data where ground-truth is available, we use reference-based metrics: Peak Signal-to-Noise Ratio (PSNR)\cite{PSNR} and the Structural Similarity Index Measure (SSIM)\cite{SSIM} and show violin plots showing the distributions of scores over the full test for a selected number of models. We report the mean of each score over the test set for all models in Appendix~\ref{sec: restoration_appendix}. For the degraded experimental set, which lacks a ground-truth, we employ Kernel Inception Distance (KID)\cite{KID} and CLIP Maximum Mean Discrepancy (CMMD)\cite{CMMD} to compare the generated image distribution against the distributions of both pristine and degraded experimental images. In all following metrics, $A(i,j)$ and $B(i,j)$ represent two $n \times m$ images from sets $P$ and $Q$ respectively, $\alpha_{net}$ and $\beta_{net}$ represent the embeddings of $A(i,j)$ and $B(i,j)$ respectively given by a model $net$ which is either the CLIP or InceptionV3 model \cite{radford2021learning, szegedy2016rethinking}.

\subsubsection*{Peak Signal-to-Noise Ratio (PSNR)}
PSNR quantifies the reconstruction quality of an image by measuring the ratio of its maximum possible power to the power of corrupting noise. Expressed in decibels (dB), a higher PSNR value corresponds to a lower level of reconstruction error.
\begin{equation}
    PSNR = 10 \cdot \log_{10}\left(\frac{\text{MAX}_I^2}{\text{MSE}}\right)
\end{equation}

where $MAX_I$ is the maximum possible pixel value, and the Mean Squared Error (MSE) between two $n\times m$ images, $I(i,j)$ and $K(i,j)$, is given by:
\begin{equation}
    MSE = \frac{1}{mn}\sum_{i=0}^{m-1}\sum_{j=0}^{n-1} [I(i,j) - K(i,j)]^2
\end{equation}

\subsubsection*{Structural Similarity Index Measure (SSIM)}
SSIM assesses perceptual image quality by quantifying the degradation of structural information. It models this degradation as a combination of three factors: luminance, contrast, and structure. The resulting index ranges from -1 to 1, where 1 signifies perfect similarity.
\begin{equation}
    SSIM(A,B) = \frac{(2\mu_A\mu_B + c_1)(2\sigma_{AB} + c_2)}{(\mu_A^2 + \mu_B^2 + c_1)(\sigma_A^2 + \sigma_B^2 + c_2)}
\end{equation}
$\mu_A$, $\mu_B$ are the means of images $A$ and $B$ respectively, $\sigma_A$, $\sigma_B$ are the standard deviations of images $A$ and $B$ respectively, $\sigma_{AB}$ is the covariance of images $A$ and $B$, and $c_1$ and $c_2$ are small constants to stabilise the division if the denominator is very small.

\subsubsection*{Kernel Inception Distance (KID)}  
KID is an alternative to the widely used Fréchet Inception Distance (FID)\cite{FID}. Like FID, it evaluates generative models by measuring the dissimilarity between feature distributions of real and generated images extracted from the InceptionV3 network \cite{szegedy2016rethinking}. However, unlike FID, which assumes these distributions are Gaussian, KID provides a non-parametric comparison using the squared Maximum Mean Discrepancy ($MMD^2$) with a polynomial kernel $k(\alpha_{inc}, \beta_{inc}) = \left(\frac{1}{d}\alpha_{inc}^\top \beta_{inc} + 1\right)^3$, where $d$ is the feature dimension. This makes KID a more robust and reliable estimator than FID, particularly on smaller datasets. A lower KID score indicates a smaller discrepancy. The MMD score is given by
\begin{equation}\label{eq:mmd}
\begin{split}
MMD^2(P,Q) ={}& E_{\alpha_{inc},\alpha_{inc}'}[k(\alpha_{inc},\alpha_{inc}')] \\
&+ E_{\beta_{inc},\beta_{inc}'}[k(\beta_{inc},\beta_{inc}')] \\
&- 2E_{\alpha_{inc} , \beta_{inc}}[k(\alpha_{inc},\beta_{inc})]
\end{split}
\end{equation}
where $E[\cdot]$ is the expectation value.

\subsubsection*{CLIP Maximum Mean Discrepancy (CMMD)}
CMMD operates similarly to KID but utilises the pre-trained Contrastive Language-Image Pre-Training (CLIP) model to embed images into a multi-modal feature space. The distance between the feature distributions of real and generated images is then quantified using the squared MMD (Eq.~\ref{eq:mmd}) with a Gaussian kernel \( k(\alpha_{clip}, \beta_{clip}) = \exp\left(-\|\alpha_{clip} - \beta_{clip}\|^2 / \sigma^2 \right) \), with the bandwidth parameter set to $\sigma=10$. As with KID, a lower CMMD score denotes a higher perceptual similarity.

\

\section{Results}
We evaluate model performance on image restoration and super-resolution (SR), using both synthetically degraded STM images and experimental degraded images of Si(001):H. 
We first present a qualitative visual assessment, as shown in Figure~\ref{fig:denoise1}, followed by a quantitative analysis.

On synthetic data, we used the reference-based PSNR and SSIM. For SR, we deliberately omit PSNR due to its known limitations in assessing perceptual quality in this task\cite{ledig2017photo}. For experimental data, we used the no-reference KID and CMMD to assess perceptual quality and distributional similarity.

We set empirical lower and upper bounds for each task to contextualise the perceptual metrics. For restoration, the bounds correspond to pristine-pristine as the lower bound and pristine-degraded as the upper bound; for super-resolution, they correspond to high-resolution-high-resolution as the lower bound and high-resolution-low-resolution as the upper bound.

Unless otherwise stated, for each architecture-size pairing, we report the best-performing configuration selected over all available number of inference steps. In the tables, the DDIM variant trained with the Fast Fourier Trasform loss is labelled as FFT. Complete tabulations for every configuration, including model family, model size, and inference step count, are provided in Appendices~\ref{sec: restoration_appendix} and \ref{sec: SR_appendix}.

\subsection*{Image Restoration}

We begin with a qualitative analysis of the models' ability to restore experimentally degraded images, for which we acquired a reliable ground-truth sequence. We first conditioned the STM tip for stable, high-resolution imaging, after which the same surface area was scanned repeatedly without repositioning. This process induced natural tip degradation over successive scans, yielding a series of images of the same region, from pristine to progressively degraded. Although minor differences persist (e.g. non-linear distortions from creep and hysteresis, or slight surface changes due to tip-sample interactions), this sequence provides a physically realistic basis for qualitatively assessing model performance.

\begin{figure}[htbp!]
    \centering
    \includegraphics[width=\columnwidth ]{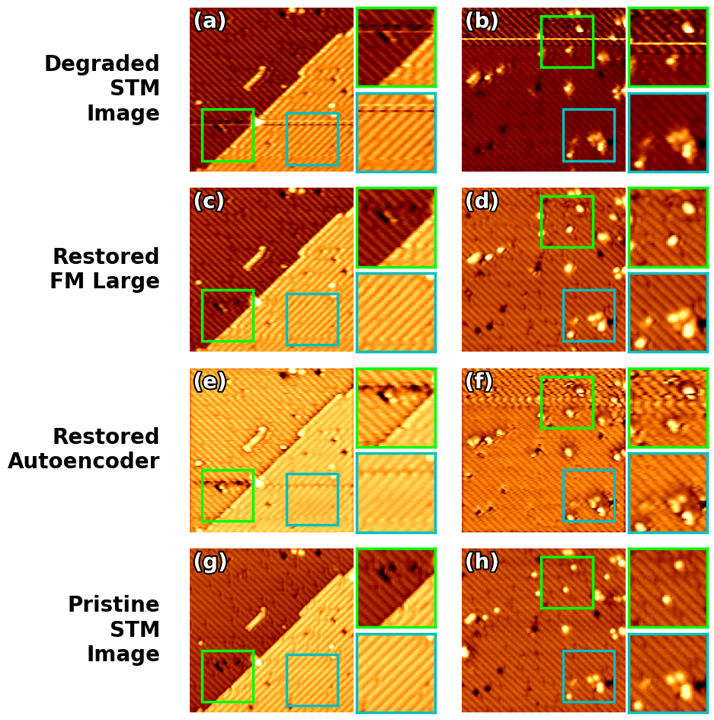}
    \caption{Qualitative results of the image restoration models on experimental STM data. The first and last rows are real experimental images (images taken at -2~V, 20~pA) of the same 25~nm $\times$ 25~nm area of Si(001):H. The middle rows show restored versions of the raw data (a, b) using two models: FM Large with two inference steps (c, d) and the Autoencoder (e, f). FM Large more effectively suppresses artefacts like scan line noise and preserves atomic detail, while the Autoencoder tends to introduce distortions such as blurriness or unnatural contrast. }\label{fig:denoise1}
\end{figure}

Figure~\ref{fig:denoise1} compares FM Large with the Autoencoder on two representative cases. The moderately degraded input in Figure~\ref{fig:denoise1}(a) exhibits both positive and negative scan line noise, as well as scan line misalignments. Both models improve scan line noise. FM Large additionally corrects the upper scan line misalignment, but not the lower one. Neither model fully removes the blurriness between the noisy scan lines; however, FM Large avoids introducing new artefacts, whereas the Autoencoder produces dark shadowing in the blurred regions, as seen in the lower inset of Figure~\ref{fig:denoise1}(e).

Figure~\ref{fig:denoise1}(b) shows a more severely degraded image, affected by a tip change, scan line noise, and a multi-tip artefact. The FM Large model restores the image effectively, removing blurriness, scan line noise, and multi-tip features. However, the restoration is not perfect, as the bright features appear slightly elongated relative to their true size. In contrast, the Autoencoder introduces additional distortions and an unnatural global contrast and fails to recover the true shapes of the surface features.

\begin{figure}[htbp!]
    \centering
    \includegraphics[width=\columnwidth ]{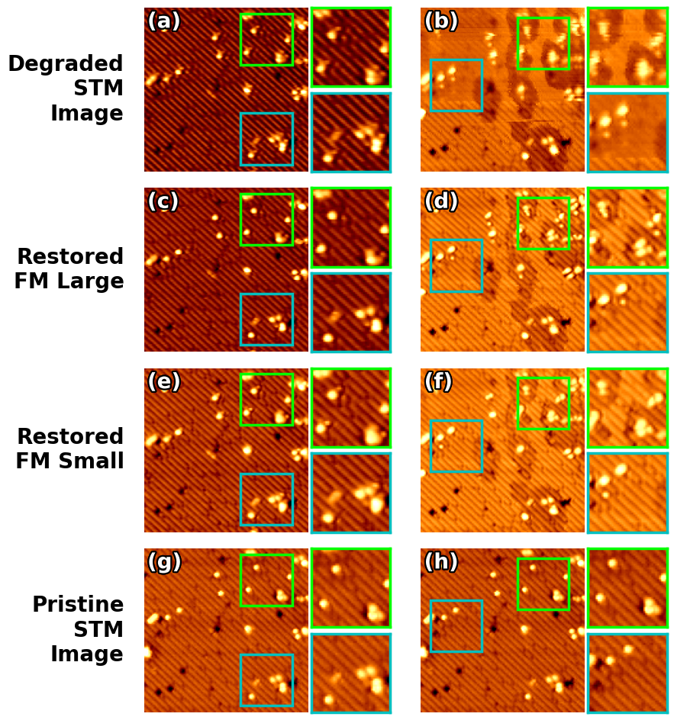}
    \caption{Qualitative results of the image restoration models on experimental STM data. The first and last rows are real experimental images (images taken at -2~V, 20~pA) of the same 25~nm $\times$ 25~nm area of Si(001):H. The middle rows show restored versions of the raw data (a, b) using two models: FM Large with two inference steps (c, d) and FM Small with two inference steps (e, f). The two restorations are similar, although FM Large has done slightly better at removing the double tip artefact in (a). Neither model succeded in restoring image (h) completely.} \label{fig:denoise2}
\end{figure}

Figure~\ref{fig:denoise2} compares FM Large with FM small. The degraded input in Figure~\ref{fig:denoise2}(a) exhibits a multi-tip. FM Large successfully removes this feature. FM Small is able to partially correct it, but not as effectively as FM Large. However, FM Small does not introduce the same artefacts (e.g. unnatural contrast) as the Autoencoder does. Figure~\ref{fig:denoise2}(b) is a more severely degraded image. The top half is extremely blurry, with scan line misalignments, a multi-tip. Additionally, the dark depression around the point defect indicates tip induced charging, which is a real surface property \cite{schofield2013quantum,labidi2015scanning}. Although the outputs of both FM Large and FM Small look realistic, the restoration is not true to the pristine STM image. The underlying lattice is effectively restored, but both models fail to restore the correct number of bright and dark features and their shapes. The second column highlights the fact that this model should not be used in all cases, particularly when changes to the image represent real changes on the surface  and or have complex physical origins, such as the tip induced charging.

\begin{table}[htbp!]
    \centering
    \caption{Image restoration: PSNR and SSIM on synthetically degraded test sets containing only a single type of degradation. Values are the mean scores averaged across all generative models to assess the relative difficulty of restoring each isolated degradation type.}
    \label{tab:noise_types_denoising}
    \begin{tabular}{lcc}
        \toprule
        \textbf{Degradation Type} & \textbf{PSNR ↑} & \textbf{SSIM ↑} \\
        \midrule
        Blunt Tip       & $32.40$ & $0.953$ \\
        Scan Lines      & $32.38$ & $0.964$ \\
        Tip Change      & $31.04$ & $0.951$ \\
        Misalignment    & $30.36$ & $0.938$ \\
        Multi-tip       & $27.18$ & $0.870$ \\
        \bottomrule
    \end{tabular}
\end{table}

When visually inspecting the figures, it is clear that the models remove certain types of degradation more effectively than others. To quantitatively assess which degradation types pose the greatest challenge, we created five separate test sets, each containing only one form of degradation. Table~\ref{tab:noise_types_denoising} reports the average PSNR and SSIM for these sets across all models. Among the different artefacts, the multi-tip dataset stands out as the most challenging degradation, with an average PSNR score more than 3 dB lower and an SSIM score 0.068 lower than any other category.

For an initial quantitative comparison, we evaluated all models on a synthetically degraded test set containing all degradation types illustrated in Figure~\ref{fig:noise}, exploring various architectures and numbers of inference steps to balance restoration quality with computational cost. 

\begin{table}[htbp!]
\setlength{\tabcolsep}{6pt}
\centering
\caption{Quantitative performance for the image restoration task, measured by mean PSNR and SSIM on the synthetic test set. Diffusion models are compared across different architectures and size against the Autoencoder baseline.}
\label{tab:psnr-ssim-denoising-main}
\begin{tabular}{llrr}
\toprule
\textbf{Model} &  \textbf{PSNR} & \textbf{SSIM} \\
\midrule
Synthetically Degraded Images & 20.00 & 0.680 \\
\midrule
Autoencoder &  18.82 & 0.787 \\
DDIM Small &  26.59 & 0.890 \\
DDIM Large (FFT) & 28.86 & 0.910 \\
FM Small & 29.36 & 0.895 \\
FM Large & 31.57 & 0.929 \\
\bottomrule
\end{tabular}
\end{table}

\begin{figure}[htbp!]
    \centering
    \includegraphics[width=\columnwidth ]{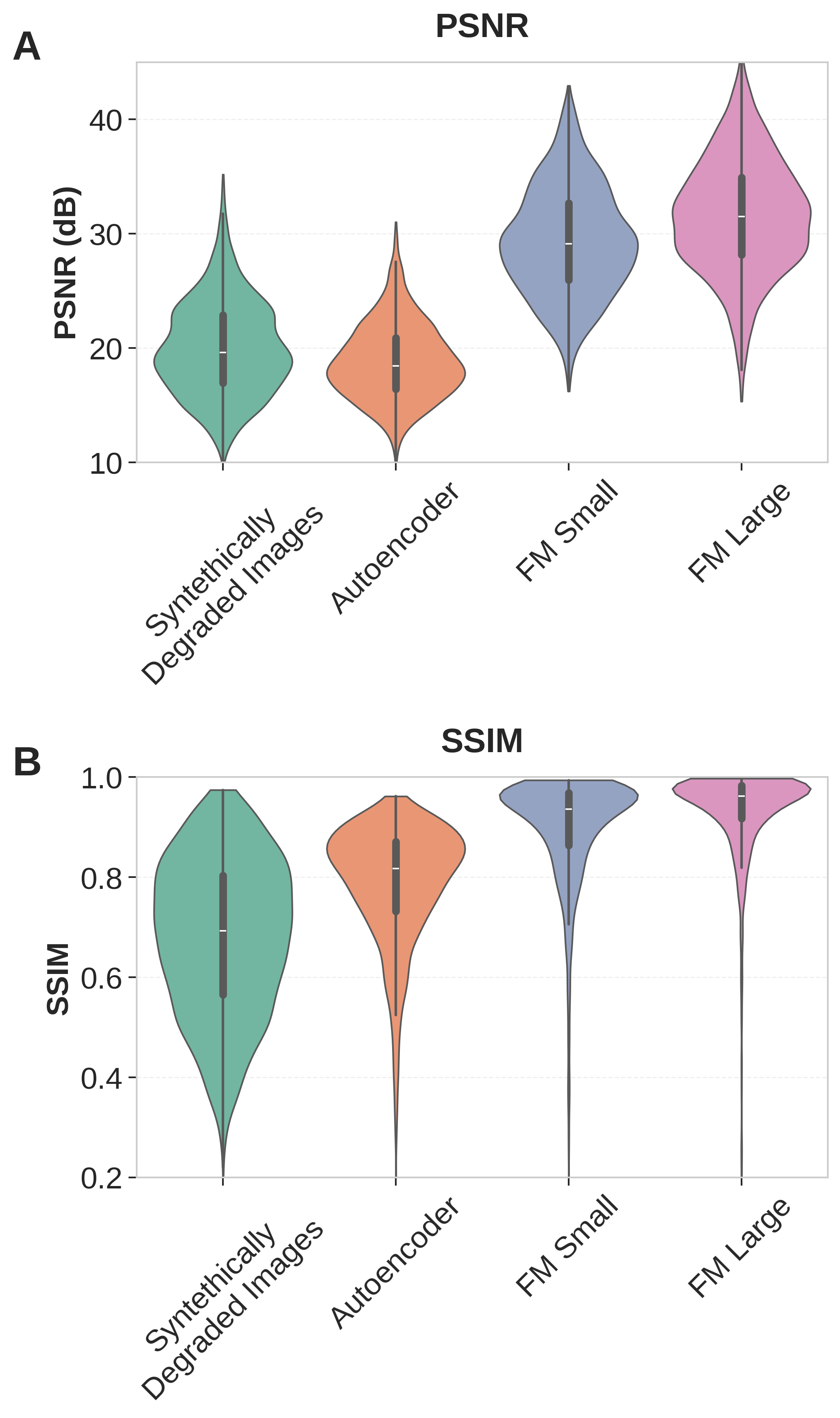}
    \caption{Quantitative evaluation of image restoration models on the synthetic test set. The violin plots show the distribution of (A) Peak Signal-to-Noise Ratio (PSNR) and (B) Structural Similarity Index Measure (SSIM) for the noisy inputs, the Autoencoder baseline, the best-performing small generative model, and the best-performing large model.}
    \label{fig:denoising_pnsr_ssim}
\end{figure}

Figure~\ref{fig:denoising_pnsr_ssim} shows the PSNR and SSIM distributions for a selection of the best performing models, comparing the best small generative model and the best large generative model against the Autoencoder baseline. The violin plots clearly illustrate that both generative models achieve substantially higher and more consistent scores than the Autoencoder. The comprehensive results for all variants tested are detailed in the appendix in Table~\ref{tab:psnr-ssim-denoising-full}.

Evaluating model performance solely on synthetically degraded data would not provide a complete picture of their ability to restore real experimental images. Therefore, in Table~\ref{tab:perceptual-denoising-main}, we report KID and CMMD scores to assess perceptual fidelity on the experimental dataset. The results confirm the superior performance of the generative models over the Autoencoder baseline. As shown in Table~\ref{tab:perceptual-denoising-main}, the Flow-Matching (FM) models generally yield the lowest (best) KID scores, indicating that their output distributions are perceptually closest to the pristine reference images. Performance on the CMMD metric is more mixed, with both top-tier FM and DDIM models achieving strong, comparable results. The Autoencoder, however, is a significant outlier with high KID and CMMD scores, placing its output distribution much closer to that of the noisy images and reflecting substantially lower perceptual quality. The complete numerical results for all models are provided in the appendix in Table~\ref{tab:perceptual-denoising-full}.

\begin{table}[htbp]
\centering
\caption{Perceptual quality of image restoration on the experimental test set, evaluated with Kernel Inception Distance (KID) and CLIP Maximum Mean Discrepancy (CMMD).The 'Pristine Images' and 'Degraded Images' rows represent the ideal target and the worst-case baseline, respectively.}
\label{tab:perceptual-denoising-main}
\begin{tabular}{lrr}
\toprule
\textbf{Model} & \textbf{KID ↓} & \textbf{CMMD ↓} \\
\midrule
Synthetically Degraded Images & 0.0807 & 0.446 \\
\midrule
Autoencoder & 0.0830 & 0.431 \\
DDIM Small & 0.0365 & 0.375 \\
DDIM Large (FFT) & 0.0397 & 0.336 \\
FM Small & 0.0331 & 0.349 \\
FM Large & 0.0357 & 0.350 \\
\midrule
Pristine Images & 0.0194 & 0.228 \\
\bottomrule
\end{tabular}
\end{table}

\subsection*{Super Resolution}

\begin{figure}[htbp]
    \centering
    \includegraphics[width=\columnwidth ]{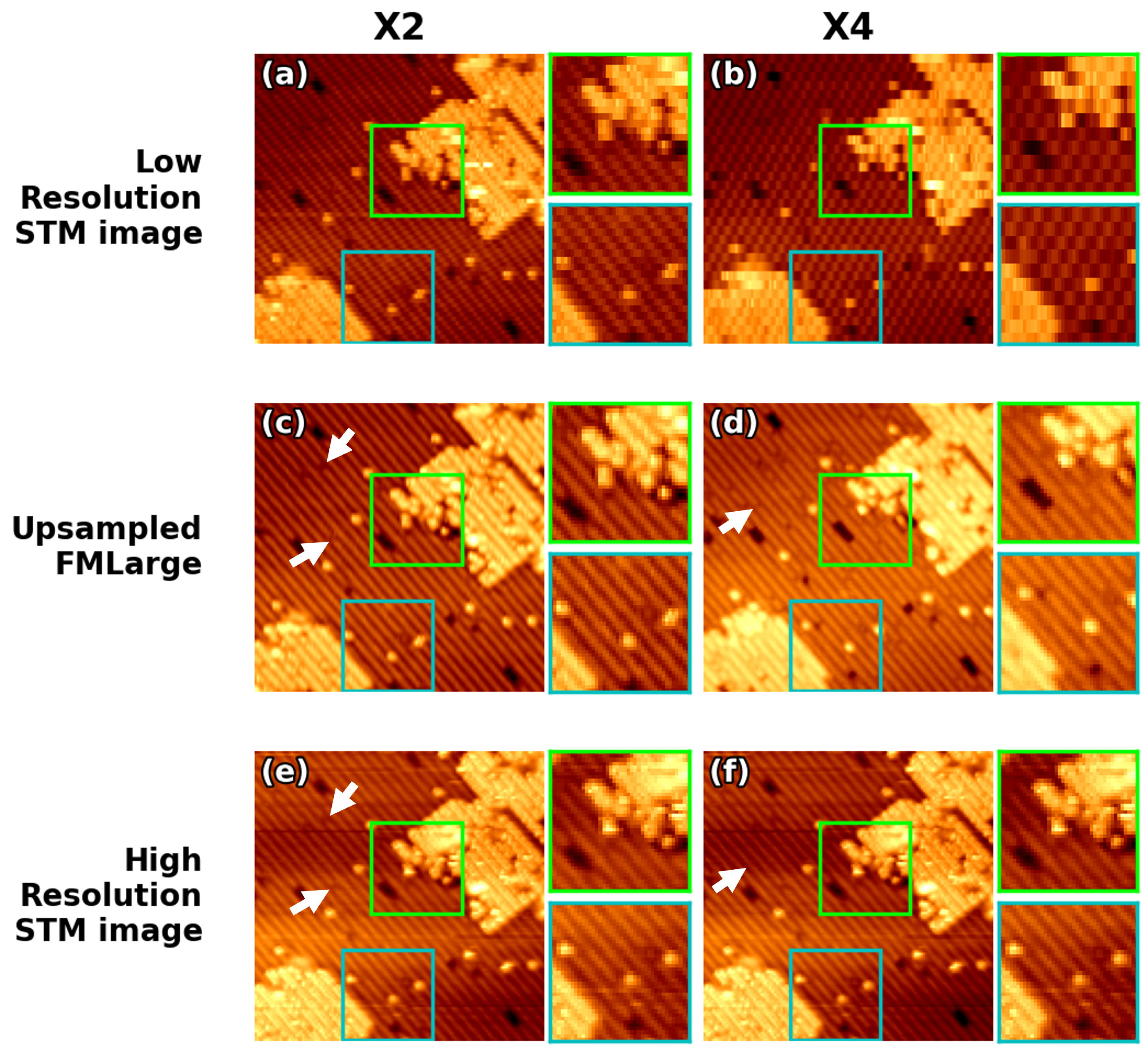}
    \caption{Qualitative results for $\times$2 (a, c, e) and $\times$4 (b, d, f) super-resolution on experimental STM images (taken at -2~V, 30~pA) of the same 25~nm $\times$ 25~nm area of Si(001):H. For each upscaling factor, the low-resolution input (top row) is reconstructed by the FM Large model (middle row) and compared against the high-resolution ground-truth image (bottom row). The model successfully restores fine atomic details, though minor discrepancies in small defects are visible in the $\times$4 case, as shown in the insets.}
    \label{fig:sr1}
\end{figure}
We train separate models for $\times$2 and $\times$4 upsampling and include realistic degradations during training to reflect experimental conditions.
Qualitative performance on experimental data is illustrated in Figure~\ref{fig:sr1}. 
For both scaling factors, the FM Large model successfully reconstructs fine atomic features, including dimer rows and surface defects, from low-resolution input.
A closer comparison of the fourfold case shows small discrepancies in the number and location of subtle single atom defects (siloxane)\cite{croshaw2020atomic}, pointed out by the white arrows in Figures~\ref{fig:sr1}(d) and ~\ref{fig:sr1}(f).
This limitation reflects the effective pixel spacing of the low-resolution input approaching the characteristic size of the defects, thereby setting a practical upper bound on recoverable detail at high magnification. Nevertheless, the results demonstrate that the super-resolution pipeline can substantially reduce acquisition time while preserving strong structural fidelity.

To assess the relative difficulty of different degradation types in the super-resolution (SR) task, we evaluated isolated artefact test sets at each scale factor, as summarised in Table~\ref{tab:noise_types_SR}. As expected, performance decreases with increasing upsampling factor, with the average SSIM dropping by 0.27 when moving from $\times$2 to $\times$4, confirming that higher magnification introduces greater reconstruction difficulty. Interestingly, in the $\times$2 task, the scan line noise and blunt tip datasets achieve SSIM values identical (to three decimal places) to those of the low-resolution-only baseline. This may indicate that the corresponding parameter distributions in the synthetic pipeline were not fully representative of these artefacts for SR training. Consistent with the image restoration task, most degradation types in the $\times$2 setting show comparable relative difficulty, with the multi-tip artefact standing out as the most challenging, scoring 0.31 lower than the next lowest case. At $\times$4, the spread in difficulty becomes more pronounced, though the multi-tip degradation again remains the most difficult, with an SSIM 0.29 below the next worst performer.

\begin{table}[htbp!]
    \centering
    \caption{Impact of different degradation types on super-resolution performance for $\times$2 and $\times$4 upscaling. Values are the mean SSIM scores, averaged across all models, on synthetic test sets containing a single artifact type. The "Low Resolution" row serves as a baseline, showing performance on images that were only downsampled without additional degradation.}
    \label{tab:noise_types_SR}
    \begin{tabular}{lrr}
        \toprule
        \textbf{Degradation Type} & \textbf{SSIM($\times$2)} & \textbf{SSIM($\times$4)} \\
        \midrule
        Low Resolution    & $0.750$  & $0.723$ \\
        \midrule
        Scan Lines      & $0.750$  & $0.703$ \\
        Blunt Tip       & $0.750$  & $0.713$ \\
        Misalignment    & $0.747$  & $0.721$ \\
        Tip Change      & $0.746$  & $0.723$ \\
        Multi-tip       & $0.717$  & $0.672$ \\
        \bottomrule
    \end{tabular}
\end{table}

Quantitative performance on the synthetic SR test set, measured by SSIM, is presented in Figure~\ref{fig:sr_pnsr_ssim} and Table~\ref{tab:psnr-ssim-sr-main}. The complete SSIM break down appears in Appendix~\ref{sec: SR_appendix} (Table~\ref{tab:ssim-sr-full}).
Across both $\times$2 and $\times$4 upscaling, the Flow-Matching models achieve the highest reconstruction fidelity, with FM Large scoring the best in the $\times$2 task (0.778) and tying with FM Small in the $\times$4 task (0.719). Notably, the Autoencoder baseline outperforms the DDIM models on this pixel-wise metric.
The violin plots in Figure~\ref{fig:sr_pnsr_ssim} reveal the full SSIM distributions and highlight this gap: both FM models are shifted upward with narrower spreads and fewer low-score tails than the Autoencoder and the low-resolution inputs, indicating higher central fidelity and consistency at both scale factors. Table~\ref{tab:ssim-sr-full} in Appendix~\ref{sec: SR_appendix} also clarifies step sensitivity and variants: FM models reach their best scores with small step counts (e.g., FM Large at 2 steps), while DDIM Large benefits modestly from the FFT loss.

\begin{table}[htbp!]
    \centering
    \caption{Quantitative performance for the super-resolution task for $\times$2 and $\times$4 upscaling, measured by mean SSIM on the synthetic test set. Diffusion models are compared across different architectures and size against the Autoencoder baseline.}
    \label{tab:psnr-ssim-sr-main}
    \begin{tabular}{lrr}
        \toprule
        \textbf{Model} & \textbf{SSIM($\times$2)} & \textbf{SSIM($\times$4)} \\
        \midrule
        Synthetically Degraded Images & 0.493 & 0.396 \\
        \midrule
        Autoencoder  & 0.747 & 0.672 \\
        DDIM Small & 0.703 & 0.661 \\
        DDIM Large (FFT) & 0.724 & 0.660 \\
        FM Small & 0.757 & 0.719 \\
        FM Large & 0.778 & 0.719 \\
        \bottomrule
    \end{tabular}
\end{table}

\begin{figure}[htbp!]
    \centering
    \includegraphics[width=\columnwidth ]{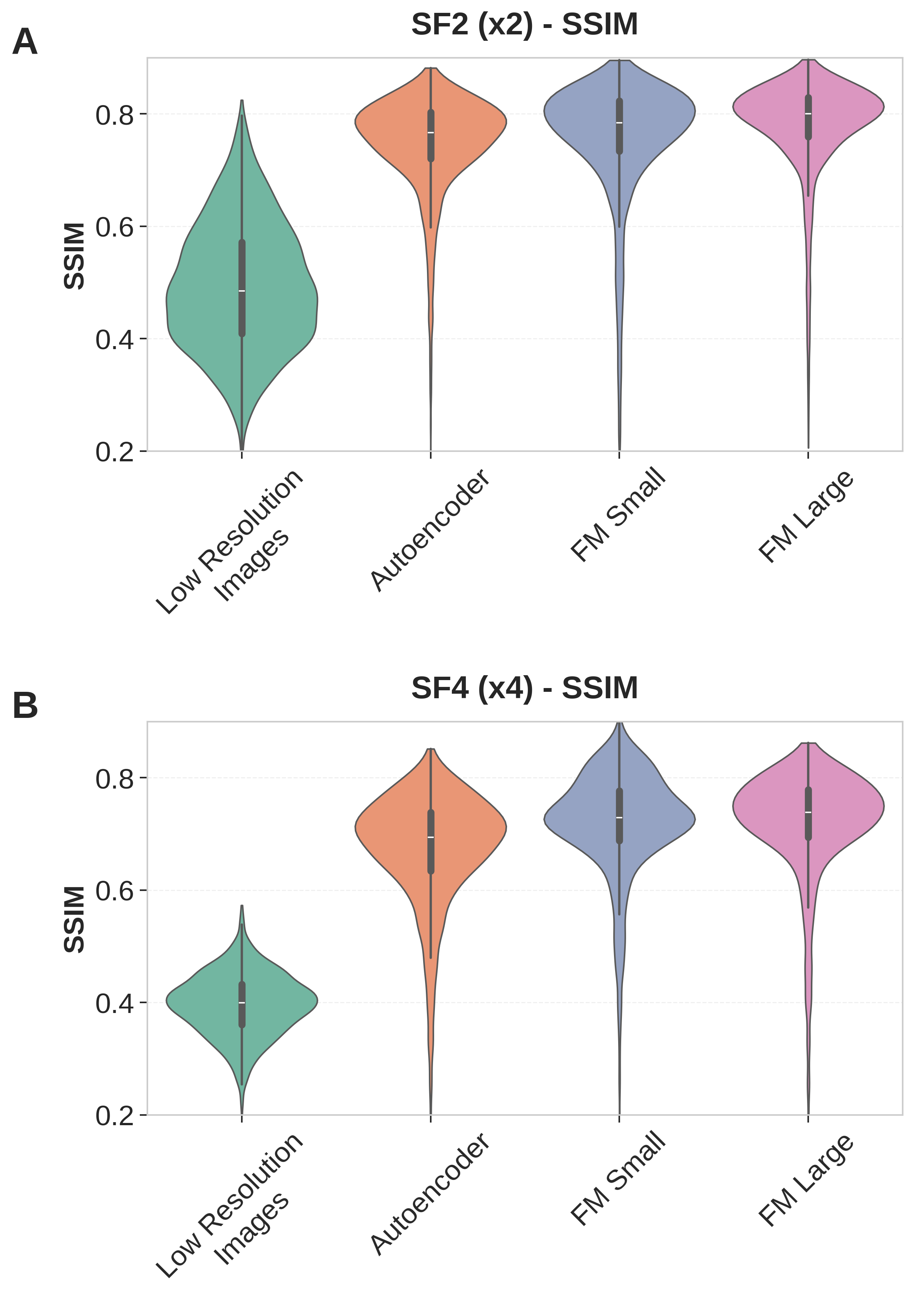}
    \caption{Quantitative evaluation of super-resolution (SR) models on the synthetic test set for different scale factors (SF). The violin plots show SSIM distributions for $\times$2 upsampling (A) and $\times$4 upsampling (b) for the Low-resolution Images, the Autoencoder baseline, the best-performing small generative model, and the best-performing large model.}
    \label{fig:sr_pnsr_ssim}
\end{figure}

The perceptual quality on the experimental SR dataset was evaluated using KID and CMMD, with results summarised in Table~\ref{tab:perceptual-sr}. These metrics reveal a trade-off between the models. For perceptual similarity (KID), the FM Small model is best in both scale factors, achieving the lowest (best) scores for both $\times$2 (0.0977) and $\times$4 (0.0983) upscaling. In contrast, for distribution alignment (CMMD), the DDIM Small model performs best on the $\times$2 task (0.361), while the FM Large and DDIM Large models are tied for the best score on the $\times$4 task (0.362). The complete results for all SR model variants are available in Appendix~\ref{sec: SR_appendix} in Tables~\ref{tab:kid-sr-full} and ~\ref{tab:cmmd-sr-full}. These breakdowns show that FM Small attains its best KID with few steps (e.g., 2 steps), whereas DDIM Large achieves its best CMMD at 10 steps and gains modestly from the FFT loss, especially at fourfold upscaling.

\begin{table}[htbp]
\centering
\caption{Perceptual quality of the super-resolution (SR) task on the experimental test set, evaluated for both $\times$2 and $\times$4 upsampling, evaluated with Kernel Inception Distance (KID) and CLIP Maximum Mean Discrepancy (CMMD). Lower scores for both metrics indicate better performance. The 'Low Resolution' and 'High Resolution' rows represent the ideal target and the worst-case baseline, respectively.}
\label{tab:perceptual-sr}
\begin{tabular}{lrrrr}
\toprule
\textbf{Model} & \textbf{KID ($\times$2)} & \textbf{KID ($\times$4)} & \textbf{CMMD ($\times$2)} & \textbf{CMMD ($\times$4)} \\
\midrule
Low Resolution & 0.1436 & 0.1782 & 0.479 & 0.601 \\
\midrule
Autoencoder & 0.1021 & 0.1275 & 0.407 & 0.392 \\
DDIM Small  & 0.1106 & 0.1359 & 0.361 & 0.414 \\
\begin{tabular}{@{}r@{}}DDIM Large \\ (FFT)  \end{tabular} & 0.1138 & 0.1020 & 0.373 & 0.362 \\
FM Small    & 0.0977 & 0.0983 & 0.401 & 0.370 \\
FM Large    & 0.1081 & 0.0993 & 0.408 & 0.362 \\
\midrule
High resolution & 0.0187 & 0.0187 & 0.229 & 0.229 \\
\bottomrule
\end{tabular}
\end{table}

\subsection*{Real Time Lab Integration}

A critical aspect of our framework is its practical utility in a standard laboratory. We therefore benchmarked the inference speed of our models on consumer-grade hardware (AMD Ryzen 5 2600 CPU; NVIDIA RTX 3060 Ti GPU) to assess their real-world performance to reconstruct a 128$\times$128 pixel image. Appendix \ref{sec: timing_appendix} Table~\ref{tab:inference-timing-full} reports inference times for all models across all variants and step counts.

Since the total inference time for the generative models scales linearly with the number of steps, we report the average time per step as a key performance metric, summarised in Table~\ref{tab:inference-timing}. This value determines the base unit of the computational cost of a reconstruction. On a standard CPU, the DDIM Small model is the most efficient, requiring only 0.10 seconds per step. The FM Small model is also highly performant at 0.21 seconds per step

\begin{table}[htbp!]
\centering
\caption{Inference speed for a 128$\times$128 image. CPU and GPU represent the average time in seconds per inference step. Speedup is the ratio of CPU to GPU time.}
\label{tab:inference-timing}
\begin{tabular}{lrrr}
\toprule
\textbf{Model} & \textbf{CPU (s)} & \textbf{GPU (s)} & \textbf{Speedup ($\times$)} \\
\midrule
Autoencoder     & 1.08 & 0.08 & 13.50 \\
DDIM Small      & 0.10 & 0.01 &  9.73 \\
DDIM Large      & 1.13 & 0.08 & 13.87 \\
FM Small        & 0.21 & 0.02 & 10.30 \\
FM Large        & 2.27 & 0.16 & 14.08 \\
\bottomrule
\end{tabular}
\end{table}

Crucially, this means that a few-step reconstruction with a small generative model can be faster than a single pass of an Autoencoder. For example, a DDIM Small model with 5 steps completes a reconstruction of a 128$\times$128 pixel image in just 0.5 seconds on a CPU, offering the advanced capabilities of a generative model at a low computational cost. This shows that relatively small diffusion models are computationally efficient enough to be integrated into experimental workflows without specialised hardware. As expected, leveraging a dedicated GPU provides a significant 10-14$\times$ speed-up, enabling near real-time processing for even the largest models.

\section{Discussion}
We have demonstrated a deep learning framework that addresses the persistent STM challenges of image degradation and slow acquisition. The strength of our approach is a physics-informed augmentation pipeline that requires only a small set of pristine experimental images, bypassing the need for costly simulations or extensive manual labelling. By considering the origins of common STM image artefacts, we were able to construct augmentation functions that reproduce these effects with sufficient realism to enable generative models trained on synthetic data to generalise to real experimental images. This demonstrates that carefully designed synthetic datasets can substitute, at least in part, for the labour-intensive curation of large experimental datasets.

Our models are capable of restoring images affected by a wide range of common STM degradations, although the analysis indicates that complex artefacts arising from multi-tip effects remain the most challenging to correct. This difficulty likely stems from several factors: when only two bright features are visible, it can be inherently ambiguous whether these represent distinct surface protrusions or a duplicated signal from a single feature; moreover, our current multi-tip augmentation model is likely too simplistic to capture the full range of physical variations present in such artefacts. The parameters of this model were empirically selected based on visual inspection of the generated data. Future work could refine this process by quantitatively tuning these parameters against experimental multi-tip datasets and perceptual quality metrics such as the CMMD. In contrast, other degradation types were removed with relative ease. For example, although a blunt tip results in some loss of high-frequency information, the models appear to have learnt the underlying Si(001):H lattice symmetry and can restore it effectively. Similarly, scan line noise possesses a distinctive structure and intensity that make it relatively straightforward for the models to identify and suppress. 

Beyond effective artefact removal, it is crucial that the models neither introduce nor eliminate genuine surface features. The acceptable level of modification ultimately depends on the specific application, but, ideally, such alterations should be minimised. Although we do not have a direct quantitative metric to assess hallucinated features, Figures~\ref{fig:denoise1} and \ref{fig:denoise2} provide qualitative evidence. In Figure~\ref{fig:denoise1}(b, f), while the restorations are not perfect (e.g., slight blurring near the repaired scan line in (b) and mild elongation of features in (f)), the key surface characteristics - such as lattice non-uniformities - are preserved, with no clear addition or removal of physical features. Only in Figure~\ref{fig:denoise2}(f) does FM Large produce a visually plausible but categorically incorrect image compared to Figure~\ref{fig:denoise2}(h). Notably, the model still retains charging effects (depressions around some bright features), despite these being absent from the training set, demonstrating a degree of generalisation and physical consistency. This outcome is desirable, as charging effects represent genuine surface phenomena rather than tip artifacts, indicating that the models preserve intrinsic surface properties rather than removing them.

Nonetheless, not all degraded images can be meaningfully restored, as the following examples illustrate. Figure~\ref{fig:denoise2}(e) represents a case where the degradation is so severe that a meaningful restoration is not possible - indeed, even an experienced STM user would find interpretation difficult. In practice, we advise that when an STM image has degraded to the point where manual interpretation would be unreliable, model-based restoration is unlikely to yield meaningful results. In such cases, tip conditioning remains necessary.

The quantitative evaluation provides further insight into the strengths and limitations of our models, but the limitations of the chosen metrics must be taken into account. The FM models consistently outperform the DDIM and Autoencoder baselines, although the difference between DDIM and FM is not as large. SSIM specifically reflects their ability to restore the finer atomic details. Notably, the difference between FM Large and FM Small is relatively modest, indicating that even the smaller models achieve good reconstruction quality, which is particularly relevant for practical laboratory implementation.
While PSNR and SSIM are useful metrics that can tell us how well single images have been restored, they do not fully capture the models' generalisation abilities as they can only be evaluated on the synthetic test set. Complementary no-reference metrics, such as KID and CMMD, are therefore essential for assessing the quality of restored experimental data. Both metrics show a considerable improvement in perceptual quality of the images when comparing them to our top and bottom baselines for all models, except the Autoencoder. However, the CMMD and KID scores disagree on the ranking of the models and on the average degree of improvement. This discrepancy likely arises from the differences in how the underlying networks, CLIP and InceptionV3, were trained.

The InceptionV3 network was trained on approximately one million images from 1000 classes, leading to embeddings that capture broad perceptual qualities such as textures and edges. The KID scores being closer to the lower bound (pristine image) than the upper bound (degraded images) suggests that the models successfully reproduce general lattice textures and surface structures characteristic of STM images i.e. the lattice background, edges etc. On the other hand, CLIP was trained on 400 million (image, text) pairs, resulting in rich embeddings that can capture higher-level semantic meaning, i.e. density, spatial arrangement, and size of features. The smaller CMMD improvement may therefore reflect differences in the density and morphology of surface features between the restored and pristine images, rather than a true perceptual degradation.

However, since both networks were trained on natural rather than STM images, their embeddings may not map the same perceptual characteristics relevant to STM data. The observed disagreement between KID and CMMD should therefore be interpreted cautiously. Regardless, all models perform similarly relative to the degraded and pristine baselines, suggesting that CMMD and KID are broadly consistent in indicating substantial improvement, but have limited sensitivity to subtle differences between models.

We further demonstrated that our generative models are effective for super-resolution, enabling up to a fourfold acceleration in image acquisition and thereby enhancing experimental throughput. Quantitative analysis (Table~\ref{tab:noise_types_SR}) indicates that, apart from severe multi-tip artefacts, most degradation types do not substantially impair SR performance beyond the inherent difficulty of upsampling from low-resolution data. However, in practice, we observe that degraded low-resolution inputs are reconstructed less accurately than pristine ones - particularly for $\times$4 upsampling. For instance, scan-line artefacts are occasionally misinterpreted as genuine surface features after upsampling. Ideally, the model would remain robust to such degradations, but at $\times$4 resolution the generative process tends to hallucinate when the input contains significant artefacts. Nevertheless, the SR framework remains valuable: it reliably doubles the resolution of degraded images and accurately reconstructs high-quality inputs at both $\times2$ and $\times4$ scales, within the physical limits imposed by pixel size.

The qualitative results highlight the intrinsic limitations of this process imposed by pixel size. As shown in Figure~\ref{fig:sr1}, features smaller than the effective pixel spacing of the low-resolution input cannot be faithfully recovered, leading to discrepancies in the reconstruction of atomic-scale defects. 

On experimental data, KID and CMMD reveal a trade-off: FM Small yields the lowest KID at twofold and fourfold, indicating strong perceptual similarity to the pristine distribution, whereas DDIM Large achieves the best CMMD at twofold and ties with FM Large at fourfold, suggesting closer alignment of feature densities and spatial statistics at moderate scaling. Taken together, SSIM and KID favour FM for texture and local structure, while CMMD highlights DDIM Large’s strength in matching global arrangement. 

Failure modes remain consistent with restoration. Multi-tip is the dominant adverse condition for SR, whereas scan line noise, misalignment, and blunt tip blur are more readily corrected. Hallucination risk is low in typical cases but increases under extreme degradation where a human expert would also hesitate; in such cases, abstention and reacquisition remain the safest course. Consequently, this pipeline is best understood not as a tool for generating ground-truth data, but as an accelerator for applications where speed is prioritised over perfect atomic fidelity, such as sample navigation or high-throughput fabrication.

Additionally, a human expert evaluation was initiated to assess the perceptual quality of restored and super-resolved images. The methodology is presented in Appendix~\ref{sec: human_eval_appendix} with initial results in Table~\ref{tab:human_eval_summary} that align with quantitative findings. Evaluation is ongoing for both denoising and super-resolution to consolidate these preliminary results.

This work complements established experimental practices like physical tip conditioning, and its data efficient nature ensures transferability to new STM systems and materials with minimal additional data.

\section{Conclusion}
This study shows that diffusion models can be trained to restore and upsample degraded STM images even with limited experimental data, by leveraging a data generation methodology that models tip artefacts. Across synthetic and experimental evaluations, FM and DDIM models consistently outperformed a conventional autoencoder on both fidelity and perceptual alignment. Qualitative analyses showed restoration of scan line noise, misalignment, and blunt tip blur, while quantitative benchmarks confirmed clear gains across reference-based and perceptual metrics. The models inference timing support near real-time deployment within standard laboratory workflows on commodity hardware.

Our objective is to assist rather than replace experimental practice. In contrast to methods that aim for simulation-quality images or to fully automate physical tip repair as Joucken et al. and Xie et al., our framework reduces the strictness of tip conditioning required for reliable surface interpretation. By broadening the range of acceptable tip conditions, it increases the yield of interpretable data. The pipeline is data-efficient and readily transferable to new material systems, with a clear path to adaptation for other scanning probe modalities such as AFM.

The study also clarifies the limits of current capabilities. Multi-tip artefacts remain the dominant failure mode for restoration, and at fourfold super-resolution small-defect counts and shapes can deviate when the effective pixel spacing approaches defect dimensions.

Taken together, the results demonstrate that generative models can enhance STM workflows in real time on accessible hardware, offering a data-efficient tool that complements existing procedures and improves experimental throughput.

\section{Experimental}
The substrate system explored here (Si(001):H) is prepared using standard procedures \cite{stock2024single}. Measurements were made using an Omicron variable temperature STM at room temperature.

\section*{Acknowledgments}
This project was financially supported by the Engineering and Physical Sciences Research Council (EPSRC) [grant numbers EP/V027700/1, and EP/W000520/1], Innovate UK [grant number UKRI/75574], and ERC AdvG ENERGYSURF. N.L.K. was partly supported by the EPSRC Centre for Doctoral Training in Advanced Characterisation of Materials [grant number EP/L015277/1] and Nanolayers Research Computing. The authors acknowledge the use of the UCL Myriad High Performance Computing Facility (Myriad@UCL), and associated support services, in the completion of this work.

Tommaso Rodani was supported by the European Union – NextGenerationEU, M4C2, within the PNRR project NFFA-DI, CUP B53C22004310006, IR0000015, having benefited from the access provided by AREA Science Park in Trieste.

The authors acknowledge the AREA Science Park super-computing platform ORFEO made available for conducting the research reported in this paper and the technical support of the Laboratory of Data Engineering staff. 

The authors wish to thank Kieran Spruce, Jamie Bragg, Ois\'in Fitzgerald, and Timothy Brown for dedicating their time and providing preliminary feedback during the expert evaluation. 

\section*{Author Contributions}
T.R. and N.L.K. contributed equally to this work.

Tommaso Rodani: Data curation (equal); Formal analysis (equal); Investigation (equal); Methodology (equal); Software (equal); Writing – original draft (equal); Writing – review \& editing (equal). Nikola L. Kolev: Data curation (equal); Formal analysis (equal); Investigation (equal); Methodology (equal); Software (equal); Writing – original draft (equal); Writing – review \& editing (equal).
Neil J. Curson: Conceptualization; Funding acquisition; Supervision; Writing – review \& editing. 
Taylor J. Z. Stock: Conceptualization (equal); Supervision (equal); Writing – review \& editing (equal). 
Alberto Cazzaniga: Conceptualization (equal); Funding acquisition (equal); Supervision (equal); Writing – review \& editing (equal).

\section*{Data availability}
The datasets used in the current study are publicly released in the Zenodo repository 10.5281/zenodo.17474268.
\section*{Code availability}
The code used to train the models and perform the analysis presented in this study is available at \href{https://github.com/RitAreaSciencePark/physics-informed-stm-restoration}{https://github.com/RitAreaSciencePark/physics-informed-stm-restoration}.
\section*{Conflict of interest}
The authors declare no conflict of interest.

\section*{References}

\bibliographystyle{unsrt}
\bibliography{bibliography}

\clearpage

\section*{Appendix}
\appendix
\subsection{STM Imaging of the Si(001):H Surface and Hydrogen Desorption Lithography} \label{sec:STM_background}

\begin{figure}[htbp!]
    \centering
    \includegraphics[width=0.45\textwidth]{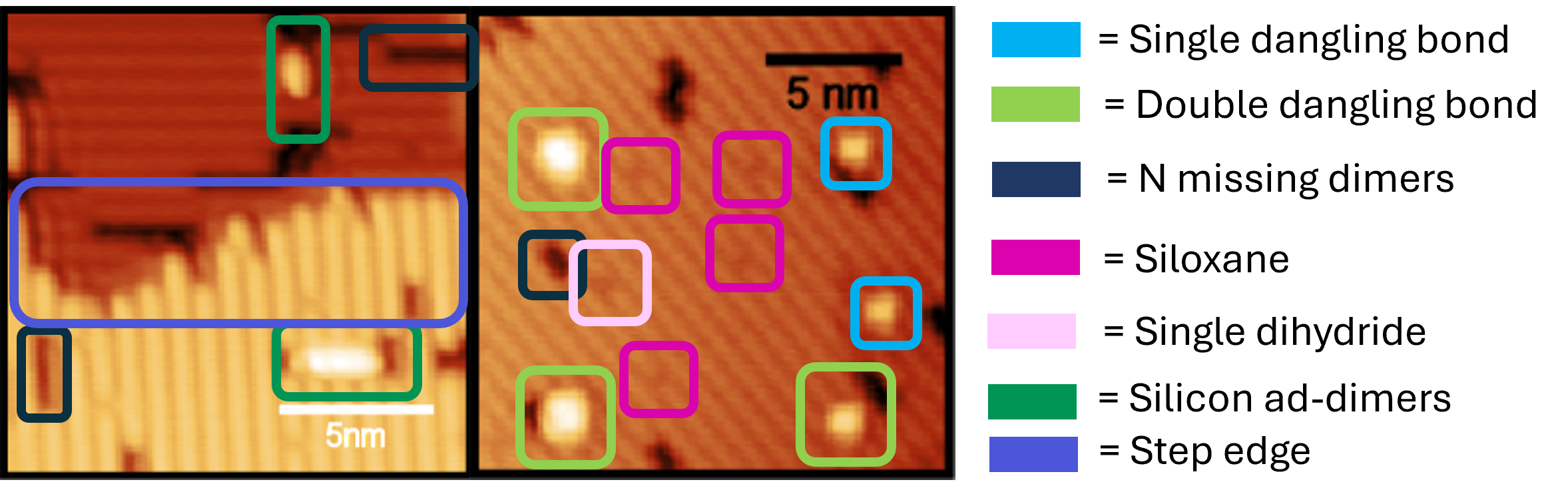}
    \caption[short]{STM images of the Si(001):H surface acquired at -2~V and 30~pA. Different features of interest are circled in the image such as step edges, missing dimers (missing silicon atoms), and dangling bonds (missing hydrogen atoms). The columns and rows that can be seen are referred to as dimer rows.}
    \label{fig:si(001)H_exp}
\end{figure}

In Figure~\ref{fig:si(001)H_exp} we show two STM images of a Si(001):H surface and indicate different features of interest with coloured boxes. The Si(001):H surface has a $2\times1$ reconstruction. Silicon dimers form in rows and these are shown running both vertically and horizontally in the left hand image and at $45^{\circ}$ in the right. In the hydrogen terminated surface, each surface silicon atom is bound to one hydrogen atom. This hydrogen monolayer is unreactive and can act as a resist for lithographic surface pattering. Using the STM in surface manipulation mode, we can selectively desorb hydrogen atoms below the tip by applying a large bias (between 3~V - 8~V) \cite{lyding1998ultrahigh}. This leaves a bare and highly reactive silicon surface. Features highlighted in pale blue and pale green in the right hand image are reactive sites where one or two hydrogen atoms are missing, exposing single or double silicon dangling bonds. After using the STM tip to pattern larger areas of these reactive dangling bond sites, we can then deterministically place dopant atoms in the surface by exposure to a precursor gas, such as arsine or phosphine, that is precision leaked into the ultra-high vacuum chamber used to prepare and measure the sample.

\subsection{Derivation of a Double Tip Artefact} \label{sec: Double_tip_derivation}

In this Appendix, we motivate the use of Equation~\ref{eqn: double_tip_eqn} by deriving a quantum-mechanical expression for the tunnelling current in the case of a double tip. The derivation follows from the Tersoff-Hamann model under the assumptions stated below. The final form is slightly modified to generate more visually realistic synthetic multi-tip images.

\begin{figure}[htbp!]
    \centering
    \includegraphics[width=0.35\textwidth]{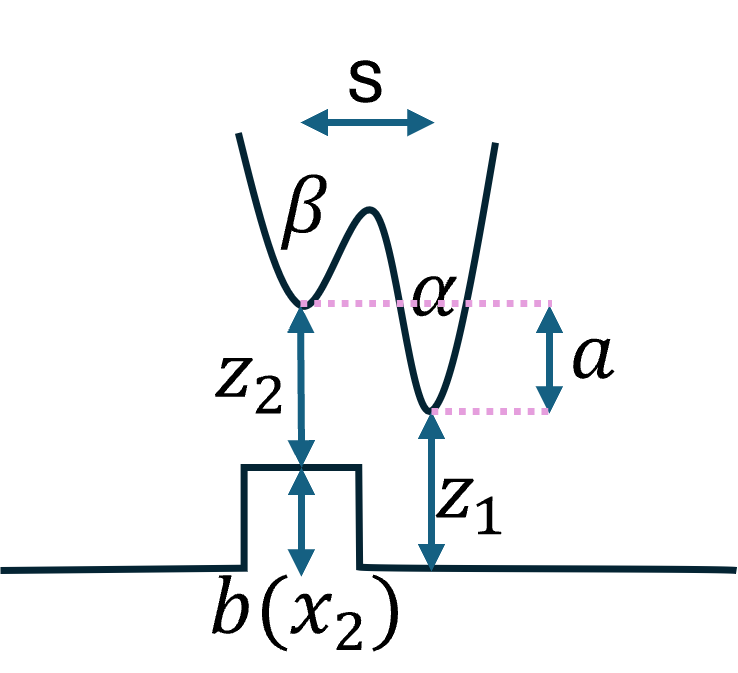}
    \caption[Illustration of a double tip.]{Illustration of a double tip. The centre of mass (CoM) of tip $\alpha$ is at $r_1$, and the CoM of tip $\beta$ is at $r_2$. Tip height difference is a. Difference in height of surface below tip $\beta$ is $b(x_2)$.}
    \label{fig:double_tip}
\end{figure}

We model the STM tip as two point sources of tunnelling current, each with a height above the surface of $z_1$ and $z_2$ for tip $\alpha$ and tip $\beta$ respectively. The distance between the two tips is $s$. Without loss of generality, we take both tips to be on the x-axis. The length difference between the two tips is $a$. The difference in height of the surface below tip $\beta$ compared to tip $\alpha$ is $b(x_2)$. 

From Straton et al. \cite{straton2020clarifying}, the total tunnelling current, $I_T$, for the double tip is given by:

\begin{equation}
    I_T = I_{\alpha} + I_{\beta} + I_{\alpha\beta} + I_{\beta\alpha}
\end{equation}

Where $I_{\alpha}$ and $I_{\beta}$ are the tunnelling currents between each tip individually and the surface i.e. electrons taking classical paths. $I_{\alpha\beta}$ and $I_{\beta\alpha}$ are interference terms, i.e. electrons taking quantum paths. 

We assume that the interference terms are negligible since the tunnelling current decays exponentially with distance, and the tip-tip separations in the augmentations are 0.2-3.1~nm. Even at the smallest separation, the extra distance travelled by the electrons to the alternate tip is significant. This assumption is consistent with Mizes et al.~\cite{mizes1987multiple}, who also neglect interference, whereas Straton et al. note that such effects depend on the separation and sample lattice.

From the Tersoff-Hamann model\cite{tersoff1985theory}, the tunnelling current for each tip individually is given by:

\begin{equation}
    I_{i} = C_i \sum_{\nu} |\psi_{\nu}(r_i)|^2 \delta(E_{\nu}-E_F) \quad i = \alpha,\beta 
\end{equation}

Where $C_i$ are constants that depend on bias, sample workfunction, density of states at the Fermi energy and tip curvature. $\psi_{\nu}(r_i)$ is the wavefunction of the surface evaluated at the centre of the tip $i$. $E_{\nu}$ is the energy of the state $\nu$ and $E_F$ is the Fermi energy of the surface. We can expand the wavefunction into its Fourier components:
\begin{equation}
    \psi_{\nu}(r_i) = \int d^2k \ a(k) e^{-\sqrt{k^2+\kappa^2} z_i} cos(kx_i)
\end{equation}
Where $a(k)$ are the Fourier coefficients of the surface wavefunction, and $\kappa = \frac{\sqrt{2m\phi}}{\hbar}$. 
The electron probability density of a single Fourier component is given by:
\begin{equation}
    |\psi(r_i)|^2 = |a(k)|^2 e^{-2\sqrt{k^2+\kappa^2} z_i} cos^2(kx)
\end{equation}

The components with $k \approx 0$ decay the slowest, and therefore dominate the tunnelling current. We can therefore approximate the wavefunction as:

\begin{equation}
    |\psi_{\nu}(r_i)|^2 \approx e^{-2\kappa z_i}
\end{equation}

Therefore, $I_T$ can be approximated as:
\begin{equation}
    I_T \approx C_{\alpha} e^{-2\kappa z_1} + C_{\beta} e^{-2\kappa z_2}
\end{equation}

From Figure~\ref{fig:double_tip}, we can see that $z_1 = z_2 - a + b(x_2)$. Therefore, we can rewrite $I_T$ as:

\begin{equation}
    I_T = C_{\alpha} e^{-2\kappa (z_2 - a + b(x_2))} + C_{\beta} e^{-2\kappa z_2}
\end{equation} 
Rearranging,
\begin{equation}
    I_T = C_{\alpha}  e^{-2\kappa z_2} \Bigl(e^{-2\kappa (b(x_2)-a)} + \frac{C_{\beta}}{C_{\alpha}} \Bigr)
\end{equation}
\begin{equation}
    I_T = \frac{C_{\alpha}}{C_{\beta}} I_{\beta} \Bigl(\frac{C_{\beta}}{C_{\alpha}} e^{-2\kappa (b(x_2)-a)} + 1 \Bigr)
\end{equation}
\begin{equation} \label{eq: current_1}
    I_{\beta} = \frac{C_{\beta} I_T}{C_{\alpha} \Bigl( 1 + \frac{C_{\beta}}{C_{\alpha}} e^{-2\kappa (b(x_2)-a)} \Bigr)}
\end{equation}

Equation~\ref{eq: current_1} is a sigmoid-like function containing several physical parameters. To apply it, we must specify $b(x_2)$, which can be approximated as the STM-measured height at $x_2$ - i.e., the displaced version of the original image used to simulate the second tip. The remaining parameters ($I_T$, $\frac{C_{\beta}}{C_{\alpha}}$, and $a$) have physical interpretations and can be sampled from appropriate distributions for each generated image:

\begin{itemize}
    \item $I_T$ - total tunnelling current (typically 20–60~pA).
    \item $a$ - tip height difference, as defined in Figure~\ref{fig:double_tip}.
    \item $\frac{C_{\beta}}{C_{\alpha}} \approx \frac{R^2_{\beta}}{R^2_{\alpha}} e^{-\kappa(R_{\beta}-R_{\alpha})}$,
    where $R_i$ is the radius of curvature of tip $i$~\cite{tersoff1985theory}.
\end{itemize}

The final image is achieved by summing $I_\beta$ and $b(x_1)$ (the height measured by the STM at $x_1$, i.e. the original image). Although our images are acquired in constant-current mode and $I_{\beta}$ represents a changing current, both constant-current and constant-height (which produces images of the current magnitude) imaging effectively map the surface density of states at $E_F - V$.

\begin{equation} \label{eqn: hx1_1}
    h(x_1) = \frac{C_{\beta} I_T}{C_{\alpha} \Bigl( 1 + \frac{C_{\beta}}{C_{\alpha}} e^{2\kappa (b(x_1-s)-a)} \Bigr)} + b(x_1)
\end{equation}

Where $h(x_1)$ is the final height measured by the STM at point $x_1$ in our synthetic image. To simplify the expression, we also define $\gamma$ such that

\begin{equation}
    e^{\gamma} = \frac{C_\beta}{C_\alpha}
\end{equation}
Therefore, Equation~\ref{eqn: hx1_1} becomes
\begin{equation} \label{hx_final}
    h(x_1) = \frac{I_Te^{\gamma}}{\Bigl( 1 + e^{-2\kappa (b(x_1-s)-\frac{\gamma}{2\kappa}-a)} \Bigr)} + b(x_1)
\end{equation}

While the parameters in Equation~\ref{hx_final} have physical meaning, using strictly realistic values produced images that did not appear visually consistent with experimental double-tip artefacts. This discrepancy likely reflects the simplicity of the model. Consequently, the parameter distributions were tuned empirically to yield more realistic image statistics, with $\kappa$ also allowed to vary to achieve more realistic results.

By comparing Equation~\ref{hx_final} to Equation~\ref{eqn: double_tip_eqn}, we see that $A=I_te^{\gamma}$, $s = \sqrt{\tilde{x}^2+\tilde{y}^2}$,  $c = \gamma  + 2\kappa a$, and $d = \frac{\sqrt{2m\phi}}{\hbar}$. Equation~\ref{eqn: double_tip_eqn} includes displacement of tip $\beta$ in the $y$ direction as well as $x$, includes a convolution term to account for variations in tip shape, and additionally has up to 4 extra tips rather than just the one.

\subsection{Image Restoration} \label{sec: restoration_appendix}

In this appendix, we present the complete quantitative results from our model evaluations on the image restoration task.

\begin{table}[htbp!]
\setlength{\tabcolsep}{6pt}
\centering
\caption{Quantitative performance for the image restoration task, measured by mean Peak Signal-to-Noise Ratio (PSNR) and Structural Similarity Index Measure (SSIM) on the synthetic test set. Models are compared across different architectures, sizes, and numbers of inference steps.}
\label{tab:psnr-ssim-denoising-full}
\begin{tabular}{lrrr}
\toprule
\textbf{Model} & \textbf{Steps} & \textbf{PSNR ↑} & \textbf{SSIM ↑} \\
\midrule
Autoencoder & 1 & 18.82 & 0.787 \\
DDIM Small & 5 & 26.44 & 0.886 \\
DDIM Small & 10 & 26.59 & 0.890 \\
DDIM Large & 5 & 26.78 & 0.893 \\
DDIM Large (FFT) & 5 & 27.55 & 0.905 \\
DDIM Large & 10 & 26.70 & 0.892 \\
DDIM Large (FFT) & 10 & 28.86 & 0.910 \\
FM Small & 2 & 29.90 & 0.889 \\
FM Small & 5 & 28.84 & 0.893\\
FM Small & 10 & 29.36 & 0.895 \\
FM Large & 2 & 31.53 & 0.929 \\
FM Large & 5 & 31.52 & 0.929 \\
FM Large & 10 & 31.57 & 0.929 \\
\bottomrule
\end{tabular}
\end{table}

\begin{table}[htbp!]
\centering
\caption{Perceptual quality of image restoration on the experimental test set, evaluated with Kernel Inception Distance (KID) and CLIP Maximum Mean Discrepancy (CMMD).The 'Pristine Images' and 'Degraded Images' rows represent the ideal target and the worst-case baseline, respectively.}
\label{tab:perceptual-denoising-full}
\begin{tabular}{lrrr}
\toprule
\textbf{Model} & \textbf{Steps} &\textbf{KID ↓} & \textbf{CMMD ↓} \\
\midrule
Degraded Images &  & 0.0807 & 0.446 \\
\midrule
Autoencoder & 1 & 0.0830 & 0.431 \\
DDIM Small & 5 & 0.0371 & 0.377 \\
DDIM Small & 10 & 0.0365 & 0.375 \\
DDIM Large & 5 & 0.0398 & 0.339 \\
DDIM Large (FFT) & 5 & 0.0397 & 0.336 \\
DDIM Large & 10 & 0.0353 & 0.387 \\
DDIM Large (FFT) & 10 & 0.0377 & 0.345 \\
FM Small & 2 & 0.0344 & 0.365 \\
FM Small & 5 & 0.0331 & 0.349 \\
FM Small & 10 & 0.0328 & 0.353 \\
FM Large & 2 & 0.0357 & 0.350 \\
FM Large & 5 & 0.0374 & 0.350 \\
FM Large & 10 & 0.0375 & 0.357 \\
\midrule
Pristine Images & & 0.0194 & 0.228 \\
\bottomrule
\end{tabular}
\end{table}


\subsection{Super Resolution} \label{sec: SR_appendix}

In this appendix, we present the complete quantitative results from our model evaluations on the super-resolution task.

\begin{table}[htbp!]
    \centering
    \caption{Quantitative performance for the super-resolution task for $\times$2 and $\times$4 upscaling, measured by SSIM on the synthetic test set. Diffusion models are compared across different architectures and size against the Autoencoder baseline.}
    \label{tab:ssim-sr-full}
    \begin{tabular}{lrrr}
        \toprule
        \textbf{Model} & \textbf{Steps} &\textbf{SSIM($\times$2)} & \textbf{SSIM($\times$4)} \\
        \midrule
        Degraded Images & & 0.493 & 0.396 \\
        \midrule
        Autoencoder & 1 & 0.747 & 0.672 \\
        DDIM Small & 5 & 0.703 & 0.661 \\
        DDIM Small & 10 & 0.698 & 0.655 \\
        DDIM Large & 5 & 0.719 & 0.649 \\
        DDIM Large (FFT) & 5 & 0.723 & 0.660 \\
        DDIM Large & 10 & 0.722 & 0.644 \\
        DDIM Large (FFT) & 10 & 0.724 & 0.656 \\
        FM Small & 2 & 0.757 & 0.702 \\
        FM Small & 5 & 0.752 & 0.692 \\
        FM Small & 10 & 0.750 & 0.719 \\   
        FM Large & 2 & 0.778 & 0.719 \\
        FM Large & 5 & 0.762 & 0.687 \\
        FM Large & 10 & 0.764 & 0.691 \\
        \bottomrule
    \end{tabular}
\end{table}

\begin{table}[htbp!]
    \centering
    \caption{Quantitative performance for the super-resolution task for $\times$2 and $\times$4 upscaling, measured by KID on the synthetic test set. Diffusion models are compared across different architectures and size against the Autoencoder baseline.}
    \label{tab:kid-sr-full}
    \begin{tabular}{lrrr}
        \toprule
        \textbf{Model} & \textbf{Steps} &\textbf{KID($\times$2)} & \textbf{KID($\times$4)} \\
        \midrule
        Degraded Images & & 0.1436 & 0.1782 \\
        \midrule
        Autoencoder & 1 & 0.1021 & 0.1275 \\
        DDIM Small & 5 & 0.1120 & 0.1471 \\
        DDIM Small & 10 & 0.1106 & 0.1359 \\
        DDIM Large & 5 & 0.1075 & 0.1072 \\
        DDIM Large (FFT) & 5 & 0.1149 & 0.1052 \\
        DDIM Large & 10 & 0.1065 & 0.1061 \\
        DDIM Large (FFT) & 10 & 0.1138 & 0.1020 \\
        FM Small & 2 & 0.0977 & 0.0983 \\
        FM Small & 5 & 0.1022 & 0.1045 \\
        FM Small & 10 & 0.1023 & 0.1117 \\
        FM Large & 2 & 0.1081 & 0.0993 \\
        FM Large & 5 & 0.1096 & 0.1258 \\
        FM Large & 10 & 0.1103 & 0.1326 \\
        \midrule
        Pristine Images & & 0.0187 & 0.0187 \\        
        \bottomrule
    \end{tabular}
\end{table}

\begin{table}[htbp!]
    \centering
    \caption{Quantitative performance for the super-resolution task for $\times$2 and $\times$4 upscaling, measured by CMMD on the synthetic test set. Diffusion models are compared across different architectures and size against the Autoencoder baseline.}
    \label{tab:cmmd-sr-full}
    \begin{tabular}{lrrr}
        \toprule
        \textbf{Model} & \textbf{Steps} &\textbf{CMMD($\times$2)} & \textbf{CMMD($\times$4)} \\
        \midrule
        Degraded Images & & 0.479 & 0.601 \\
        \midrule
        Autoencoder & 1 & 0.407 & 0.392 \\
        DDIM Small & 5 & 0.395 & 0.419 \\
        DDIM Small & 10 & 0.392 & 0.414 \\
        DDIM Large & 5 & 0.359 & 0.384 \\
        DDIM Large (FFT) & 5 & 0.379 & 0.370 \\
        DDIM Large & 10 & 0.361 & 0.385 \\
        DDIM Large (FFT) & 10 & 0.373 & 0.362 \\
        FM Small & 2 & 0.401 & 0.370 \\
        FM Small & 5 & 0.409 & 0.373 \\
        FM Small & 10 & 0.418 & 0.390 \\
        FM Large & 2 & 0.408 & 0.362 \\   
        FM Large & 5 & 0.419 & 0.390 \\
        FM Large & 10 & 0.425 & 0.396 \\
        \midrule
        Pristine Images & & 0.229 & 0.229 \\       
        \bottomrule
    \end{tabular}
\end{table}

\clearpage

\subsection{Real Time Lab Integration}\label{sec: timing_appendix}

In this appendix, we present the complete results from our models inference timing evaluations.

\begin{table}[htbp!]
  \centering
  \caption{Inference times for reconstructing a 128$\times$128 pixel image on consumer-grade hardware.  }
  \label{tab:inference-timing-full}
  \begin{tabular}{lrrr}
    \toprule
    \textbf{Model} & \textbf{Steps} & \textbf{CPU (s)} & \textbf{GPU (s)}   \\ 
        \midrule
        Autoencoder & 1 & 1.08 & 0.08 \\
        DDIM Small & 5 & 0.50 & 0.05 \\
        DDIM Small & 10 & 1.01 & 0.10 \\
        DDIM Large & 5 & 5.68 & 0.41 \\
        DDIM Large & 10 & 11.35 & 0.82  \\
        FM Small & 2 & 0.41 & 0.04 \\
        FM Small & 5 & 1.05 & 0.10 \\
        FM Small & 10 & 2.05 & 0.20 \\
        FM Large & 2 & 4.61 & 0.32 \\   
        FM Large & 5 & 11.10 & 0.81 \\
        FM Large & 10 & 22.98 & 1.61  \\
        \bottomrule
      \end{tabular}
    \end{table}

\subsection{Human Experts Evaluation}\label{sec: human_eval_appendix}

A panel of five experienced STM users began evaluating outputs from three models selected to balance performance and efficiency: FM Small, DDIM Large, and DDIM Large FFT. The evaluation set comprised 66 experimentally degraded images of 128×128 pixels, along with 33 low-resolution images at 128×64 pixels and 33 at 128×32 pixels. Each image was restored or upsampled by the three models. The inclusion of the FFT loss variant also allowed us to evaluate its contribution to perceptual quality. 

Experts ranked the anonymized model outputs and assigned a quality rating out of five. Preliminary results from 137 annotations show a clear preference for FM Large, which achieved the highest mean quality rating (3.00/5) and best mean rank (1.87), being ranked first in 40\% of evaluations.
Conversely, DDIM Small was rated lowest, receiving the last rank 40.1\% of the time with the lowest mean rating (2.63), while DDIM Large FFT consistently performed between the two. A summary of these findings is presented in Table \ref{tab:human_eval_summary}.

Human evaluation is ongoing for both denoising and super-resolution tasks to consolidate preliminary results and ensure robust validation of models performance.

\begin{table}[htbp]
\centering
\caption{Summary of human expert evaluation for denoising. Ratings (1–5) and ranks (1–3) are mean values; higher ratings and lower ranks indicate better performance. The percentage of first place rankings is also shown.}
\label{tab:human_eval_summary}
\begin{tabular}{lrrr}
\toprule
\textbf{Model} & \textbf{Rating ↑} & \textbf{Rank ↓} & \textbf{First Rank (\%) ↑} \\
\midrule
FM Large & 3.00 & 1.87 & 40.0\%  \\
DDIM Large (FFT) & 2.77 & 2.01 & 31.6\% \\
DDIM Small & 2.63 & 2.12 & 28.5\%  \\
\bottomrule
\end{tabular}
\end{table}

\clearpage

\end{document}